\documentclass{article} 
\usepackage{iclr2026_conference}
\usepackage{times}
\usepackage{url}
\usepackage{algorithm}
\usepackage{caption}
\usepackage{pifont}
\usepackage{algpseudocode}
\usepackage{mathrsfs}
\usepackage{lipsum} 
\usepackage[normalem]{ulem}
\useunder{\uline}{\ul}{}
\usepackage{wrapfig}
\usepackage{float}
\usepackage{multirow}
\usepackage[colorlinks=true,
            linkcolor=RoyalBlue,
            citecolor=orange,
            filecolor=magenta,
            urlcolor=cyan
            ]{hyperref} 
\usepackage{adjustbox}
\usepackage{graphicx}
\usepackage{enumitem}

\usepackage{amsmath}
\usepackage{amsfonts}  
\usepackage{amssymb} 
\usepackage[normalem]{ulem}
\usepackage{dutchcal} 
\usepackage{tipx} 

\let\ul\relax
\usepackage{soul} 
\usepackage{xcolor} 
\usepackage{colortbl}

\usepackage{cleveref}
\usepackage{lipsum} 
\usepackage[normalem]{ulem}
\useunder{\uline}{\ul}{}
\usepackage{tcolorbox}
\usepackage{subcaption}
\usepackage{array}
\usepackage{booktabs}
\usepackage{longtable}
\usepackage{tabularx}

\usepackage{tikz} 
\usetikzlibrary{arrows.meta,positioning,fit,calc,decorations.pathreplacing}

\usepackage[table]{xcolor}
\usepackage{fontawesome5} 
\definecolor{schaingreen}{RGB}{0,128,0}

\usepackage[acronym]{glossaries}
\makeglossaries

\newglossarystyle{acronym4col}{%
  \setglossarystyle{long}%
  {%
    \setlength{\arrayrulewidth}{0.4pt}%
    \renewcommand{\arraystretch}{1.15}%
    \begin{longtable}{p{1cm}|p{4.5cm}|%
      p{\dimexpr\linewidth-(1cm+4.5cm+1.5cm)-6\tabcolsep\relax}|%
      p{1.5cm}}%
  }%
  {\end{longtable}}%
}

\newacronym[description={General term for systems that perform tasks requiring human-like intelligence}]{AI}{AI}{Artificial Intelligence}
\newacronym[description={Alzheimer’s is a brain disease that slowly damages memory and thinking, so everyday tasks and recognizing people become harder over time.}]{AD}{AD}{Alzheimer's disease}
\newacronym[description={Neurofibrillary tangles are twisted clumps of a protein called tau that pile up inside brain cells, jam their internal “highways,” and help kill the cells—contributing to memory loss in Alzheimer’s.}]{NFT}{NFT}{Neurofibrillary tangles}
\newacronym[description={Magnetic Resonance Imaging (MRI) is a medical imaging technique that uses strong magnets and radio waves to create detailed pictures of the inside of the body without using harmful radiation.}]{MRI}{MRI}{Magnetic Resonance Imaging}
\newacronym[description={Positron Emission Tomography (PET) is a medical imaging technique that uses tiny amounts of radioactive substances to track how organs and tissues work inside the body, creating detailed 3D pictures of their activity.}]{PET}{PET}{Positron Emission Tomography}
\newacronym[description={Cerebrospinal fluid (CSF) is a clear, watery liquid that cushions and protects the brain and spinal cord while also helping to remove waste and deliver nutrients.}]{CSF}{CSF}{Cerebrospinal fluid}
\newacronym[description={Mild Cognitive Impairment (MCI) is a condition where a person has noticeable memory or thinking problems greater than expected for their age, but not severe enough to significantly interfere with daily life or independent functioning.}]{MCI}{MCI}{Mild Cognitive Impairment}
\newacronym[description={Medial Temporal Atrophy means the shrinking of memory-related brain structures (like the hippocampus) in the inner temporal lobes, often seen in aging and Alzheimer’s disease.}]{MTA}{MTA}{Medial Temporal Atrophy}
\newacronym[description={Global Cortical Atrophy is the widespread shrinking of the brain’s outer layer (the cortex), often linked to aging or diseases like Alzheimer’s, which can affect memory, thinking, and behavior.}]{GCA}{GCA}{Global Cortical Atrophy}
\newacronym[description={Frontotemporal dementia (FTD) is a brain disorder where the nerve cells in the frontal and temporal lobes slowly waste away, leading to early changes in personality, behavior, language, and decision-making rather than memory loss (which is more typical of Alzheimer’s).}]{FTD}{FTD}{Frontotemporal Dementia}
\newacronym[description={Dementia with Lewy bodies (DLB) is a brain disorder where abnormal protein clumps (Lewy bodies) build up in nerve cells, causing a mix of memory problems, movement difficulties (like Parkinson’s), and vivid visual hallucinations.}]{DLB}{DLB}{Dementia with Lewy bodies}
\newacronym[description={The Koedam score is a visual rating scale (0–3) used on brain MRI to measure how much the parietal cortex has shrunk, helping to detect Alzheimer’s disease and other dementias.}]{Koedam}{Koedam}{Koedam score}
\newacronym[description={Beta-amyloid is a sticky protein fragment that, in Alzheimer’s disease, clumps together between brain cells to form plaques that disrupt communication and damage neurons.}]{ABeta}{A$\beta$}{Beta-amyloid}
\newacronym[description={Apolipoprotein E (APOE) is a gene that makes a protein helping transport fats in the brain, and its $\epsilon$4 version greatly increases the risk of developing Alzheimer’s disease by making brain cells more vulnerable to damage and less able to clear toxic proteins.}]{APOE}{APOE}{Apolipoprotein E}
\newacronym[description={Diffusion Tensor Imaging is an MRI technique that maps how water moves along brain fibers, helping detect early damage to the brain’s white matter connections before major memory loss or shrinkage becomes visible.}]{DTI}{DTI}{Diffusion Tensor Imaging}
\newacronym[description={A Region of Interest (ROI) in medical imaging is simply the specific part of an image—like a tumor, organ, or lesion—that researchers mark and analyze more closely because it’s the area most relevant to diagnosis or study.
}]{ROI}{ROI}{Region of Interest}
\newacronym[description={-}]{LLM}{LLM}{Large Language Model}
\newacronym[description={-}]{NLP}{NLP}{Natural Language Processing}
\newacronym[description={-}]{RNN}{RNN}{Recurrent Neural Network}
\newacronym[description={-}]{LSTM}{LSTM}{Long Short-term Memory Network}
\newacronym[description={-}]{SLM}{SLM}{Statistical Language Model}
\newacronym[description={-}]{LM}{LM}{Language Model}
\newacronym[description={-}]{PLM}{PLM}{Pretrained Language Model}
\newacronym[description={-}]{NLM}{NLM}{Neural Language Model}
\newacronym[description={-}]{GRU}{GRU}{Gated Recurrent Unit}
\newacronym[description={-}]{RLHF}{RLHF}{Reinforcement Learning with Human Feedback}
\newacronym[description={-}]{FFN}{FFN}{Feedforward Neural Network}
\newacronym[description={-}]{GPT}{GPT}{Generative Pretrained Transformer}
\newacronym[description={-}]{RL}{RL}{Reinforcement Learning}
\newacronym[description={-}]{CoT}{CoT}{Chain-of-Thought}
\newacronym[description={-}]{SFT}{SFT}{Supervised Fine-tuning}
\newacronym[description={-}]{GQA}{GQA}{Grouped-Query Attention}
\newacronym[description={-}]{SWA}{SWA}{Sliding Window Attention}
\newacronym[description={-}]{MoE}{MoE}{Mixture of Experts}
\newacronym[description={-}]{QA}{QA}{Question Answering}
\newacronym[description={-}]{ASR}{ASR}{Automatic Speech Recognition}
\newacronym[description={-}]{RLAIF}{RLAIF}{Reinforcement Learning from AI Feedback}
\newacronym[description={-}]{OOV}{OOV}{Out-of-Vocabulary}
\newacronym[description={-}]{BPE}{BPE}{Byte Pair Encoding}
\newacronym[description={-}]{BERT}{BERT}{Bidirectional Encoder Representations from Transformers}
\newacronym[description={-}]{APE}{APE}{Absolute Positional Embeddings}
\newacronym[description={-}]{RPE}{RPE}{Relative Positional Embeddings}
\newacronym[description={-}]{RoPE}{RoPE}{Rotary Positional Embeddings}
\newacronym[description={-}]{SSL}{SSL}{Self-Supervised Learning}
\newacronym[description={-}]{NLI}{NLI}{Natural Language Inference}
\newacronym[description={-}]{NLG}{NLG}{Natural Language Generation}
\newacronym[description={-}]{PEFT}{PEFT}{Parameter-Efficient Fine-tuning}
\newacronym[description={-}]{LoRA}{LoRA}{Low-Rank Adaptation}
\newacronym[description={-}]{KD}{KD}{Knowledge Distillation}
\newacronym[description={-}]{QLoRA}{QLoRA}{Quantized Low-Rank Adaptation}
\newacronym[description={-}]{VQA}{VQA}{Visual Question Answering}
\newacronym[description={-}]{MLLM}{MLLM}{Multimodal Large Language Model}
\newacronym[description={-}]{VLM}{VLM}{Vision Language Model}
\newacronym[description={-}]{HITL}{HITL}{Human-In-The-Loop}
\newacronym[description={-}]{CNN}{CNN}{Convolutional Neural Network}
\newacronym[description={-}]{ViT}{ViT}{Vision Transformer}
\newacronym[description={Computed Tomography (CT) is a medical imaging technique that uses X-rays taken from many angles and computer processing to create detailed cross-sectional pictures of the inside of the body.}]{CT}{CT}{Computed Tomography}
\newacronym[description={-}]{RAG}{RAG}{Retrieval-augmented Generation}
\newacronym[description={-}]{SV-CoT}{SV-CoT}{Structured Visual Chain-of-Thought}

\definecolor{RoyalBlue}{RGB}{65, 105, 225} 
\definecolor{custom_light_blue}{rgb}{0.85, 0.95, 1}
\definecolor{custom_light_pink}{rgb}{1, 0.85, 0.85}
\definecolor{custom_light_green}{rgb}{0.85, 0.98, 0.80}
\sethlcolor{custom_light_green}

\definecolor{IconImage}{RGB}{30,144,255}   
\definecolor{IconVideo}{RGB}{255,99,71}    
\definecolor{IconText}{RGB}{255,165,0}     
\definecolor{IconAudio}{RGB}{138,43,226}   
\definecolor{lightpeach}{RGB}{250, 190, 160}

\definecolor{headerblue}{RGB}{230,240,250}
\definecolor{headerorange}{RGB}{255, 235, 215} 

\DeclareRobustCommand{\SChain}{%
  \begingroup\normalfont
  \raisebox{-0.2em}{%
    \includegraphics[height=1.2em]{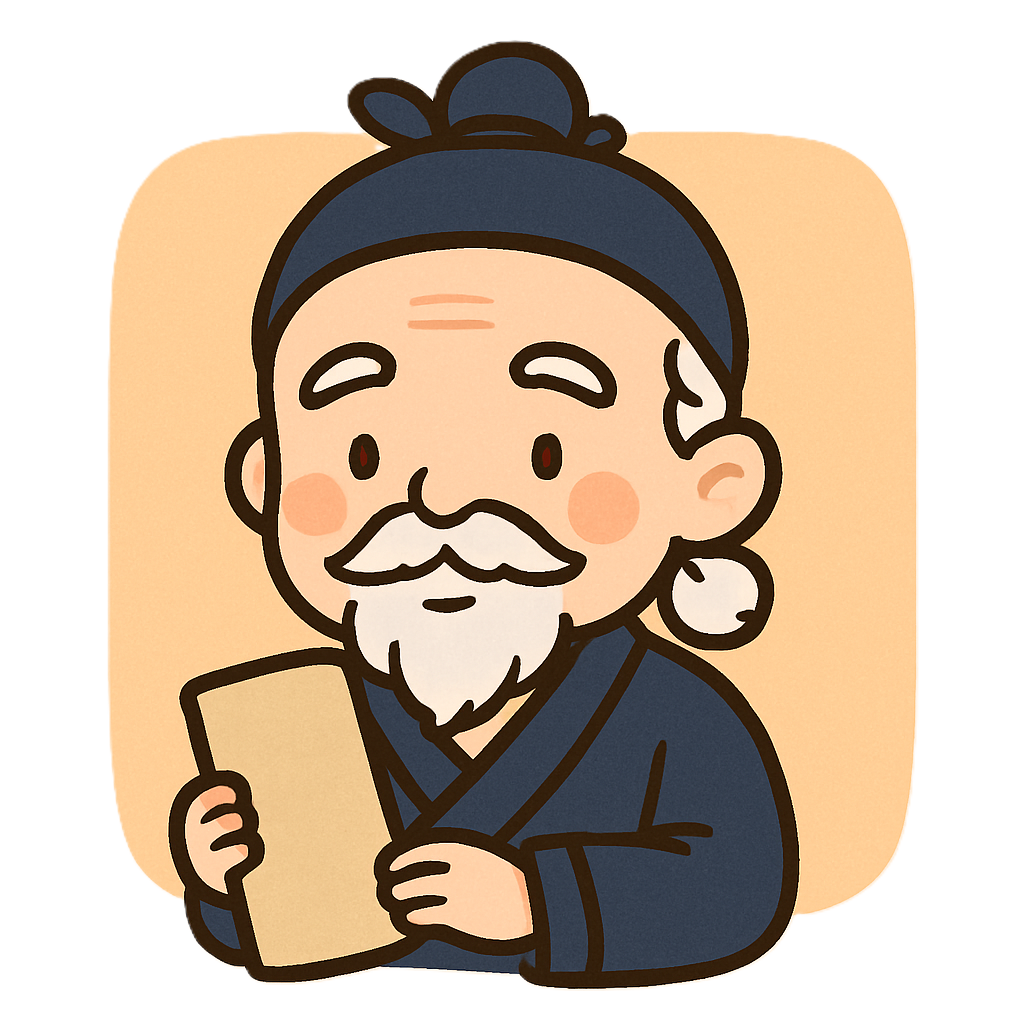}%
  }%
  \kern 0.0em 
  \textbf{\textcolor{green!60!black}{S-Chain}}%
  \endgroup
}
\definecolor{schaingreen}{RGB}{0,128,0}
\title{\SChain: Structured Visual Chain-of-Thought for Medicine}

\author{\textbf{Khai Le-Duc} \textsuperscript{$*$\,1,2} \quad \textbf{Duy M. H. Nguyen} \textsuperscript{$*$\,3,4,24} \quad \textbf{Phuong T. H. Trinh} \textsuperscript{$*$\,5} 
\quad
\textbf{Tien-Phat Nguyen} \textsuperscript{$*$\,6}
\\
\textbf{Nghiem T. Diep} \textsuperscript{$**$\,3}
\quad
\textbf{An Ngo} \textsuperscript{$**$\,7}
\quad
\textbf{Tung Vu} \textsuperscript{$**$\,8} 
\quad\textbf{Trinh Vuong} \textsuperscript{9}
\quad
\textbf{Anh-Tien Nguyen}  \textsuperscript{10,11} \\
\textbf{Mau Nguyen} \textsuperscript{12}
\quad
\textbf{Van Trung Hoang} \textsuperscript{13}
\quad
\textbf{Khai-Nguyen Nguyen} \textsuperscript{14} 
\quad \textbf{Hy Nguyen} \textsuperscript{15} 
\ \textbf{Chris Ngo} \textsuperscript{2} \\
\textbf{Anji Liu} \textsuperscript{16}
\quad \textbf{Nhat Ho} \textsuperscript{17}
\quad \textbf{Anne-Christin Hauschild} \textsuperscript{11}
\quad \textbf{Khanh Xuan Nguyen} \textsuperscript{18} \\
\textbf{Thanh Nguyen-Tang} \textsuperscript{19} 
\quad \textbf{Pengtao Xie} \textsuperscript{20,21}
\quad \textbf{Daniel Sonntag} \textsuperscript{3,22} \\
\textbf{James Zou} \textsuperscript{23} 
\quad \textbf{Mathias Niepert} \textsuperscript{4,24}
\quad \textbf{Anh Totti Nguyen} \textsuperscript{25}
\\\\
\noindent
\textsuperscript{1}\,University of Toronto, Canada\   
\textsuperscript{2}\,Knovel Engineering Lab, Singapore\\
\textsuperscript{3}\,German Research Centre for Artificial Intelligence
\textsuperscript{4}\,University of Stuttgart, Germany\\
\textsuperscript{5}\,Chonnam National University, South Korea
\textsuperscript{6}\,Singapore University of Technology and Design\\
\textsuperscript{7}\,Bucknell University, USA
\textsuperscript{8}\,Concordia University, Canada
\textsuperscript{9}\,Korea University\\
\textsuperscript{10}\,Justus Liebig University Giessen, Germany
\textsuperscript{11}\,University Medical Center G\"ottingen, Germany\\
\textsuperscript{12}\,Japan Advanced Institute of Science and Technology
\textsuperscript{13}\,Hue University, Vietnam\\
\textsuperscript{14}\,College of William \& Mary, USA
\textsuperscript{15}\,Deakin University, Australia\\
\textsuperscript{16}\,National University of Singapore
\textsuperscript{17}\,University of Texas at Austin, USA\\
\textsuperscript{18}\,University of California, Berkeley, USA
\textsuperscript{19}\,New Jersey Institute of Technology, USA\\
\textsuperscript{20}\,University of California San Diego, USA,
\textsuperscript{21}\,MBZUAI, UAE \\
\textsuperscript{22}\,Oldenburg University, Germany
\textsuperscript{23}\,Stanford University, USA\\
\textsuperscript{24}\,Max Planck Research School for Intelligent Systems (IMPRS-IS), Germany\\
\textsuperscript{25\,}Auburn University, USA
\\[3pt]
*Co-first authors; order randomized \quad **Co-second authors\\
\faEnvelope \enspace \texttt{duckhai.le@mail.utoronto.ca},\\
\hspace{1.2em} \texttt{ho\_minh\_duy.nguyen@dfki.de, anhnguyen@auburn.edu}\\
\faGithubSquare \enspace 
\href{https://github.com/leduckhai/S-Chain}{\textbf{\color{green!60!black}{S-Chain}}}
}

\iclrfinalcopy
\begin{document}
\vspace{-0.2in}
\maketitle
\vspace{-0.2in}
\begin{abstract}
\vspace{-0.1in}

Faithful reasoning in medical vision–language models (VLMs) requires not only accurate predictions but also transparent alignment between textual rationales and visual evidence. While Chain-of-Thought (CoT) prompting has shown promise in medical visual question answering (VQA), no large-scale expert-level dataset has captured stepwise reasoning with precise visual grounding. We introduce \textsc{S-Chain}, the first large-scale dataset of 12,000 expert-annotated medical images with bounding boxes and structured visual CoT (SV-CoT), explicitly linking visual regions to reasoning steps. The dataset further supports 16 languages, totaling over 700k VQA pairs for broad multilingual applicability. Using \textsc{S-Chain}, we benchmark state-of-the-art medical VLMs (ExGra-Med,  LLaVA-Med) and general-purpose VLMs (Qwen2.5-VL, InternVL2.5), showing that SV-CoT supervision significantly improves interpretability, grounding fidelity, and robustness. Beyond benchmarking, we study its synergy with retrieval-augmented generation, revealing how domain knowledge and visual grounding interact during autoregressive reasoning. Finally, we propose a new mechanism that strengthens the alignment between visual evidence and reasoning, improving both reliability and efficiency. S-Chain establishes a new benchmark for grounded medical reasoning and paves the way toward more trustworthy and explainable medical VLMs.
\end{abstract}
\vspace{-0.1in}
\addtocontents{toc}{\protect\setcounter{tocdepth}{-1}}
\section{Introduction}


\vspace{-0.1in}
\glspl{LLM} and \glspl{VLM} have shown strong capabilities in problem solving, planning, and decision making by learning deductive and inductive reasoning from large-scale data. A key driver is \gls{CoT} reasoning, which breaks complex tasks into step-by-step inferences before reaching a final answer. This paradigm improves performance across domains, from arithmetic and commonsense reasoning in \gls{LLM} \citep{wei2022chain,kojima2022large} to \gls{VQA} and multimodal reasoning in \gls{VLM} \citep{zhang2023multimodal,chen2024m}. By externalizing their reasoning process, \gls{CoT} not only boosts accuracy but also adds interpretability, making them especially promising for high-stakes fields like healthcare.

Despite recent progress, training models with strong \gls{CoT} reasoning still demands large amounts of annotated data, as models must learn to align intermediate reasoning steps with input evidence \citep{zelikman2022star,wang2022self}. In general \gls{NLP}, such supervision can be scaled through crowdsourcing or distillation \citep{magister2022teaching,ho2022large}, but in medicine, it is far more costly: annotations must be expert-verified, multimodal, and clinically valid \citep{moor2023foundation,huang2024refer}. Beyond this, medical reasoning requires visual grounding, i.e., explicitly linking reasoning steps to \gls{ROI}, which adds substantial complexity. As a result, large-scale expert datasets with grounded \gls{CoT} remain scarce, limiting the training and evaluation of trustworthy medical \glspl{VLM}.


To mitigate the high cost of expert annotation, recent work has explored auto-generation of \gls{CoT} data for \gls{VLM}  reasoning. For example, MC-CoT \citep{wei2024mc} leverages modular pipelines where \glspl{LLM} generate reasoning steps that are loosely aligned with multimodal inputs in zero-shot settings, while MedCoT \citep{liu2024medcot} introduces hierarchical expert verification to refine automatically produced rationales. Similarly, large medical \gls{VQA} datasets such as PMC-VQA \citep{zhang2023pmc} rely on template-based or synthetic \gls{QA} generation to scale supervision. While such approaches improve data availability, their effectiveness is limited for clinical reasoning due to two key issues: (i) auto-generated \glspl{CoT} often lack structure, providing free-text explanations without explicit correspondence to specific image regions, which weakens visual grounding; and (ii) they are prone to factual mistakes and hallucinations, frequently introducing redundant or clinically irrelevant content that is difficult to filter out \citep{gu2024medvh,cheng2025comt}. These limitations highlight the need for high-quality, structured, and expert-grounded \gls{CoT} annotations in the medical domain.

To address these challenges, we propose a new expert-annotated dataset that provides visually grounded \glspl{CoT} explicitly linking step-by-step reasoning to visual evidence, which we term \gls{SV-CoT}. Our dataset contains 12,000 medical images with bounding-box annotations of \gls{ROI}, paired with structured rationales that are decomposed into four clinically meaningful stages: (i) object localization, (ii) image captioning, (iii) multiple-choice reasoning, and (iv) image classification. Unlike auto-generated \glspl{CoT}, each rationale is carefully annotated and verified by medical experts, ensuring both factual accuracy and strong correspondence between reasoning steps and visual features. To enhance accessibility and global applicability, the dataset further supports \textbf{16 languages}, resulting in over \textbf{700,000 \gls{QA} pairs}. By combining structured reasoning, explicit grounding, multilingual coverage, and expert verification, this resource overcomes the key limitations of existing synthetic \gls{CoT} approaches and establishes a reliable foundation for training and benchmarking medical \glspl{VLM}.

With this dataset in place, we systematically investigate its impact on the performance of multiple model families, including both domain-specific medical \glspl{VLM} (e.g., ExGra-Med \citep{nguyen2025exgra},  LLaVA-Med \citep{li2023llava}) and general-purpose \glspl{VLM} (e.g., Qwen2.5-VL \citep{wang2024qwen2}, InternVL2.5 \citep{chen2024internvl}), and compare them against baselines trained with synthetic \glspl{CoT} generated by GPT-4.1. Beyond standard evaluation, we further assess the integration of our \gls{SV-CoT} supervision with \gls{RAG} \citep{zhao2025medrag,zheng2025miriad}, examining how external domain-specific knowledge interacts with structured reasoning and visual grounding. A key focus of our analysis is the faithfulness of \gls{CoT} reasoning and grounding during autoregressive training, where we uncover important discrepancies between textual reasoning steps and the visual evidence they reference. These findings motivate the development of new learning strategies that explicitly reinforce the correlation between grounded visual cues and \gls{CoT} reasoning, leading to more reliable, interpretable, and clinically trustworthy medical \glspl{VLM}.

In summary, we make the key contributions as:
\begin{itemize}
\vspace{-0.1in}
    \item \textbf{Dataset innovation}: We build the first large-scale dataset, \SChain, that couples 12k medical images with expert-verified bounding-box annotations and visually grounded reasoning traces, extended to 700k multilingual \gls{QA} pairs across 16 languages, structured into a four-stage reasoning pipeline to enhance clarity and consistency.
    \item \textbf{Extensive evaluation}: We conduct a broad comparative study of specialized medical \glspl{VLM} and general-purpose \glspl{VLM}, against baselines using GPT-4.1–generated rationales, highlighting the distinctive gains from expert-grounded supervision.
    \item \textbf{Analytical insights}: We examine how structured visual chain-of-thought reasoning interacts with RAG and probe the faithfulness of \gls{CoT} alignment with visual grounding during autoregressive training, from which we derive some insights for new learning strategies to tightly couple visual evidence and reasoning.
\end{itemize}


\begin{figure}[t]
    \centering
    \vspace{-0.1in}
    \resizebox{0.9\linewidth}{!}{
        \includegraphics{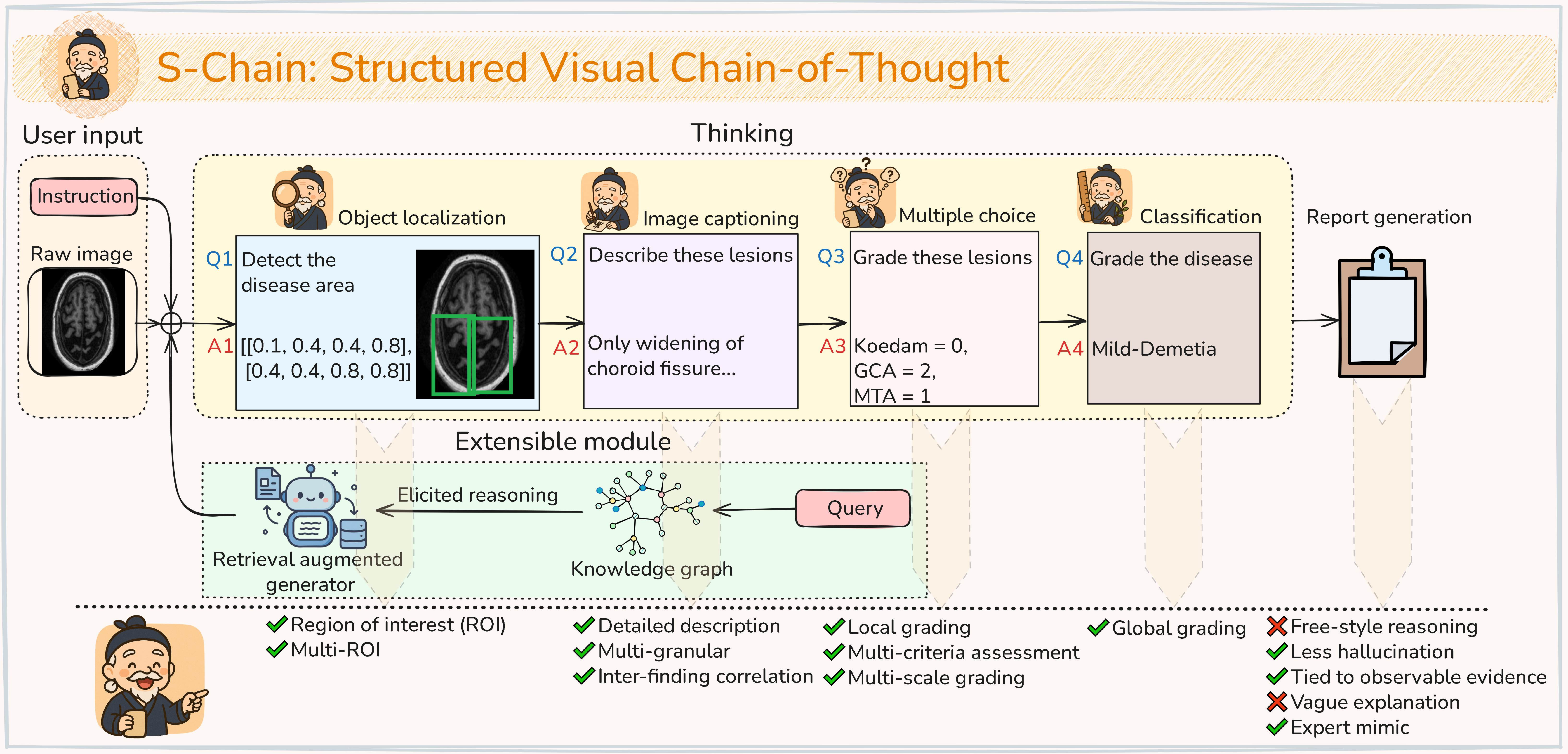}
    }
    \caption{\textbf{Overview of the \SChain\ dataset with \gls{SV-CoT} annotations}. Each image is paired with (Q1) \gls{ROI} localization via bounding boxes, (Q2) lesion descriptions, and (Q3) lesion grading using standardized scales (e.g., Koedam, GCA, MTA). These stepwise annotations ground reasoning in visual evidence, enabling interpretable and reliable medical \gls{VQA}.}
    \vspace{-0.1in}
    \label{fig:main_pipeline}
\end{figure}

\vspace{-0.15in}
\section{Problem Formulation and Key Challenges}
\vspace{-0.1in}
We study the problem of grounded medical \gls{VQA}, where the input is a medical image (e.g., a \gls{MRI} slice) together with a clinically relevant question, and the output is not only a final diagnostic answer but also a \gls{SV-CoT} that traces the reasoning process back to specific \glspl{ROI} in the image (Figure \ref{fig:main_pipeline}). In particular, the model has to (i) first identify and localize abnormalities or relevant anatomical structures with bounding boxes, (ii) then provide stepwise reasoning that links visual observations with clinical knowledge, and (iii) finally generate an interpretable answer, such as the disease type or its severity. We term this task \gls{SV-CoT}, where models must align visual-spatial cues with clinical reasoning to produce interpretable answers. Rather than giving only a final prediction, \gls{SV-CoT} forces the model to provide stepwise rationales linked to specific image regions, thereby reducing hallucinations and enabling transparent, trustworthy decision-making.

\paragraph{Prior Works.} Recent advances in medical \glspl{VLM}, such as ExGra-Med \citep{nguyen2025exgra},  LLaVA-Med \citep{li2023llavamed}, MedGemma \citep{sellergren2025medgemma}, and LLaVA-Tri \citep{xie2025medtrinity}, have primarily focused on scaling both model architectures and pre-training corpora to improve accuracy on \gls{VQA} tasks. These approaches demonstrate that larger model capacity and broader pre-training data can indeed yield stronger overall performance across diverse clinical benchmarks. Yet, despite these gains, such models remain \textit{black boxes} \citep{borys2023explainable, alsaad2024multimodal}, producing answers without revealing the clinical reasoning behind them. In practice, valid decisions require systematic analysis of markers (e.g., hippocampal shrinkage, sulcal widening, cortical thinning) and standardized scoring with \texttt{Scheltens}, \texttt{Pasquier}, or \texttt{Koedam} scales. Without reasoning chains that explicitly ground predictions in these features, models cannot provide the transparency essential for trustworthy diagnostic verification.

To enhance interpretability, several recent efforts have explored incorporating \gls{CoT} reasoning into medical \gls{AI} systems. Datasets such as MedCoT \citep{liu2024medcot}, MedThink \citep{gai2025medthink}, ReasonMed \citep{sun2025reasonmed}, and the Human-Verified Clinical Reasoning Dataset (HVCR) \citep{ding2025building} provide additional reasoning traces that improve performance and enable models to output rationales alongside predictions. However, these resources are \textit{restricted to textual \glspl{CoT}}, without linking reasoning steps to the underlying visual evidence in medical images. Other directions, such as V2T-CoT \citep{wang2025v2t}, Med-GRIT-270k \citep{huang2024refer}, and MedTrinity-25M \citep{xie2025medtrinity}, take a step further by pairing reasoning with visual grounding. Yet these datasets are largely generated \textit{using GPT-4.1–based synthetic rationales} built upon existing image–text pairs, which introduces risks of hallucination and factual errors (Figure \ref{fig:gpt_cot_quality}). Such issues are especially concerning in the medical domain, where unreliable grounding boxes or \gls{AI}-generated explanations and diagnoses may lead to misleading conclusions or inappropriate clinical guidance \citep{godinho2010clinical, shin2022dementia, monfared2024prevalence}.


In contrast, \textbf{S-Chain} introduces a dataset that directly addresses these limitations by providing expert-validated \gls{SV-CoT} for 12,000 medical images. Unlike prior synthetic or text-only resources, our dataset ensures faithful alignment between reasoning steps and visual evidence through expert-drawn bounding boxes and clinically verified rationales. Furthermore, with support for 16 languages and over 700,000 high-quality \gls{QA} pairs, it uniquely combines scale, multilinguality, and expert validation, establishing a diverse foundation for trustworthy, visually grounded reasoning in medical \glspl{VLM}. Table~\ref{tab:cot_datasets} presents an overall comparison of S-Chain with prior works in the medical domain, while Table~\ref{tab:cot_data_compare} (Appendix) extends this comparison to general-domain visual \gls{CoT} datasets.

\begin{table}[!hbt]
\centering
\vspace{-0.1in}
\caption{\textbf{Comparison of recent \textit{medical} reasoning datasets with \gls{CoT}.}}
\vspace{-0.1in}
\label{tab:cot_datasets}
\scalebox{0.7}{
\renewcommand{\arraystretch}{1.1}
\setlength{\tabcolsep}{6pt}
\begin{tabular}{l p{3.2cm} p{3.0cm} p{3.2cm} p{2.8cm} p{2.0cm}}
\toprule
\rowcolor{headerorange}
\textbf{Dataset} & \textbf{Size / Scale} & \textbf{CoT / Reasoning} & \textbf{Visual Ground.} & \textbf{Expert Involve.} & \textbf{Multiling.} \\
\midrule
MedCoT (2024) & Extends Med-VQA (VQA-RAD, SLAKE, PathVQA) & Human-verified CoTs & \textcolor{red}{\ding{55}} & \textcolor{green!60!black}{\ding{51}} Hierarchical verification & \textcolor{red}{\ding{55}} \\

MedThink (2025) & Extensions to 3 VQA sets & Decision-making rationales & \textcolor{red}{\ding{55}} & \textcolor{green!60!black}{\ding{51}} Semi-auto + human pass-through & \textcolor{red}{\ding{55}} \\

ReasonMed (2025) & 370k reasoning samples & Multi-step reasoning paths & \textcolor{red}{\ding{55}} & \textcolor{green!60!black}{\ding{51}} Multi-agent validation & \textcolor{red}{\ding{55}} \\

HVCR (2025) & 31k QA pairs & Expert-verified CoTs & \textcolor{red}{\ding{55}} & \textcolor{green!60!black}{\ding{51}} & \textcolor{red}{\ding{55}} \\

\midrule
V2T-CoT (2025) & $\sim$39k examples & GPT-generated CoTs & \textcolor{green!60!black}{\ding{51}} Partial (region attention) & \textcolor{red}{\ding{55}} (No experts) & \textcolor{red}{\ding{55}} \\

Med-GRIT-270k (2024) & 270k QA pairs & GPT-generated CoTs & \textcolor{green!60!black}{\ding{51}} Segmentation masks + region refs & \textcolor{red}{\ding{55}} (No experts) & \textcolor{red}{\ding{55}} \\

MedTrinity-25M (2024) & 25M ROI-description triplets, 10 modalities & Partial: descriptive text & \textcolor{green!60!black}{\ding{51}} ROI annotations & \textcolor{green!60!black}{\ding{51}} Expert validation ($\sim$1k subset) & \textcolor{red}{\ding{55}} \\ 
\midrule
\textbf{\SChain\ (Ours, 2025)} & \textbf{12k images / 700k QA pairs} & \textbf{Expert-verified \glspl{SV-CoT}} & \textbf{\textcolor{green!60!black}{\ding{51}} Bounding boxes (ROI links)} & \textbf{\textcolor{green!60!black}{\ding{51}} Full expert annotation (12k images)} & \textbf{\textcolor{green!60!black}{\ding{51}} (16 langs.)} \\
\bottomrule
\end{tabular}}
\end{table}

\begin{figure}[!hbt]
    \centering
     \resizebox{0.82\linewidth}{!}{
    \includegraphics[width=0.5\linewidth]{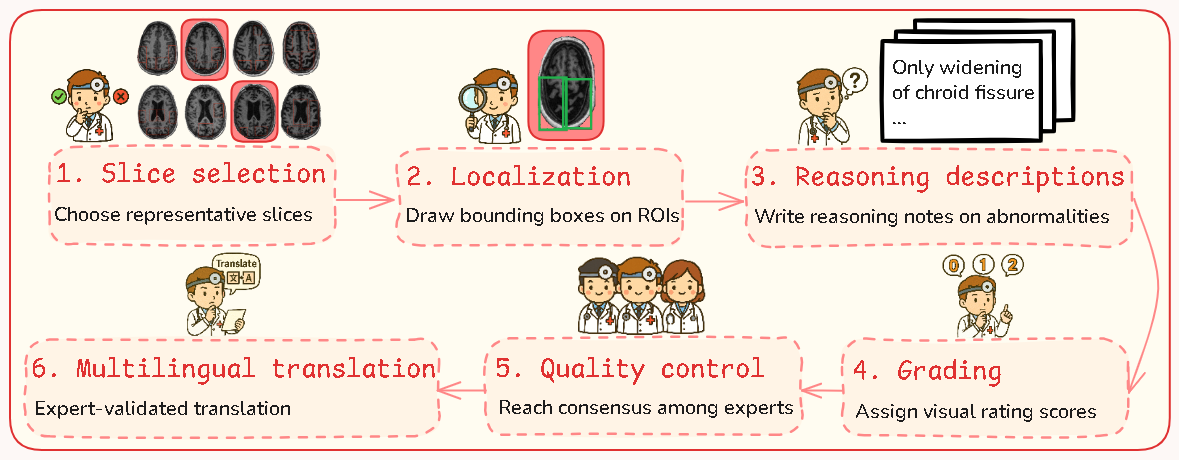}}
    \vspace{-0.05in}
    \caption{\textbf{Annotation pipeline}. Experts first select representative 2D slices from \gls{MRI} volumes (1), then localize \glspl{ROI} with bounding boxes (2). Abnormalities are described through structured reasoning notes (3) and graded using standardized visual rating scales (4). Annotations undergo expert consensus for quality control (5), and finally, all reasoning steps are translated into several languages with expert validation (6), yielding a multilingual, expert-grounded dataset. (See Appendix Section \ref{sec:dataset_examples} for some dataset examples, e.g. Figure \ref{fig:datasetexamples_english_nondementia}).}
    \vspace{-0.2in}
    \label{fig:annotation_pipeline}
\end{figure}
\section{S-Chain Dataset}
\vspace{-0.15in}
\subsection{Structured Visual Chain-of-Thought Data}
\vspace{-0.1in}
Our dataset targets the task of \gls{SV-CoT} reasoning for medical \gls{VQA}. Each example goes beyond the usual image–question–final answer prediction format by following a four-step reasoning (Figure \ref{fig:main_pipeline}) flow that mirrors clinical practice: \textbf{(Q1)} \texttt{Object localization}: bounding boxes highlight \glspl{ROI}; \textbf{(Q2)} \texttt{Lesion description}: textual explanations describe visible abnormalities (e.g., hippocampal shrinkage, sulcal widening); \textbf{(Q3)} \texttt{Lesion grading}:  findings are scored with standardized scales such as Scheltens, Pasquier, or Koedam; and \textbf{(Q4)} \texttt{Disease classification}: reasoning steps are predicted into a final diagnostic label (e.g., mild dementia). This structure tightly links visual evidence with reasoning, helping models move from black-box predictions to transparent, clinically grounded decision-making. 

\subsection{Data Collection}
\vspace{-0.1in}
We use the publicly available \gls{MRI} data from the OASIS: Cross-Sectional Alzheimer’s Disease Dataset \citep{marcus2007open}, released under the Apache 2.0 license (see Appendix Section \ref{sec:copyrights}). The dataset contains 3D brain \gls{MRI} volumes from 461 patients, accompanied by metadata including demographic information and Clinical Dementia Rating (CDR) scores. We collect patients' data that are categorized into three diagnostic groups: Non-Dementia, Mild-Dementia, and Moderate-Dementia, \textbf{with annotations provided at the volume level}. 
\vspace{-0.1in}
\subsection{Data Annotation Process}
\vspace{-0.1in}
The annotation process was conducted by three trained doctors from different institutions, working independently before consensus review. Since the OASIS dataset provides only volume-level labels, our experts first selected representative 2D slices from each 3D \gls{MRI} volume to highlight anatomical structures and pathological changes most relevant to \gls{AD}’s progression (e.g., hippocampal shrinkage, ventricular widening). On these slices, \glspl{ROI} were localized with bounding boxes, described through short reasoning notes, and graded using standardized visual rating scales. Final annotations required consensus among experts to ensure reliability. To broaden accessibility, all \gls{QA} pairs were extended into 16 languages by certified professional linguists (minimum C1 level) with basic medical training.
Figure~\ref{fig:annotation_pipeline} provides an overview of the pipeline, with stepwise details in Appendix~\ref{sec:data_anotation}. In total, constructing S-Chain required about \textbf{2100 hours of expert labor}.
\vspace{-0.1in}
\subsection{Data Statistics}
\vspace{-0.1in}

Through this process, we curated a dataset of ~12,000 expert-annotated medical images with \gls{SV-CoT}, complemented by 700k \gls{QA} pairs in 16 languages (English, German, French, Chinese, Japanese, Arabic, etc). This resource supports the development of medical \glspl{VLM} that are both multilingual and clinically reliable. As shown in Table~\ref{tab:data_stats_main}, the dataset covers 64 patients with non-overlapping train/test splits. Importantly, the test set mirrors real-world dementia cohorts (36\% Non-Dementia, 27\% Mild, 36\% Moderate) as reported in clinical studies \citep{shin2022dementia, monfared2024prevalence}, avoiding the artificially balanced splits common in \gls{AI} research and ensuring clinically meaningful evaluation.

\begin{table}[!hbt]
\centering
\vspace{-0.1in}
\resizebox{0.9\textwidth}{!}{%
\begin{tabular}{l|cccc|cc|cccc}
\toprule
\rowcolor{headerorange}
\multirow{2}{*}{} 
& \multicolumn{4}{c|}{\textbf{\#Images}} 
& \multicolumn{2}{c|}{\textbf{\#QA pairs}} 
& \multicolumn{4}{c}{\textbf{\#Patients}} \\ 
\cline{2-11} 
\rowcolor{headerorange!80!white}
& \textbf{Non} & \textbf{Mild} & \textbf{Mod} & \textbf{All} 
& \textbf{English} & \textbf{All} 
& \textbf{Non} & \textbf{Mild} & \textbf{Mod} & \textbf{All*} \\ 
\midrule
\textbf{Train} & 4,628 & 4,755 & 1,400 & 10,783 & 43,132 & 690,112 & 24 & 27 & 8 & 55 \\
\textbf{Test}  & 562 & 420 & 560 & 1,542 & 6,168 & 98,688  & 3 & 3 & 5 & 9 \\ 
\midrule
\textbf{\SChain} & 5,190 & 5,175 & 1,960 & 12,325 & 49,300 & 788,800 & 27 & 30 & 13 & 64 \\ 
\bottomrule
\end{tabular}%
}
\caption{\textbf{Statistics of S-Chain dataset}.
(*) A patient may show different labels across slices (e.g., Non-Dementia (Non) in one slice, Mild-Dementia (Mild) in another, or Moderate-Dementia (Mod) elsewhere). No overlapping of patients between train and test sets.}
\vspace{-0.15in}
\label{tab:data_stats_main}
\end{table}

\subsection{Learning SV-CoT via Supervised Fine-Tuning}
\label{subsec:train_sft}
\vspace{-0.1in}
To train medical \glspl{VLM} on \gls{SV-CoT}, we adopt an \textbf{autoregressive \gls{SFT}} strategy. 
Given an input image $I$ and a text prompt corresponding to the final question $Q_4$ (disease classification), 
the model is trained to sequentially generate multi-granularity outputs aligned with clinical reasoning steps. 
Formally, the model learns a distribution:
\setlength{\abovedisplayskip}{2pt}
\setlength{\belowdisplayskip}{2pt}
\begin{equation}
P(Y \mid I, Q_4) = \prod_{t=1}^T P(y_t \mid I, Q_4, y_{<t}),
\end{equation}
where the output sequence $Y = (Y_1, Y_2, Y_3, Y_4)$ corresponds to the structured reasoning stages: 
$Y_1$ = bounding box coordinates of \glspl{ROI} (textual form), 
$Y_2$ = lesion descriptions grounded in these regions, 
$Y_3$ = lesion grading using standardized scales, 
and $Y_4$ = the final diagnostic label. Note that the procedural questions (i.e., $Q_1$, $Q_2$, and $Q_3$) are embedded in the corresponding output sequences (i.e., $Y_1$, $Y_2$, and $Y_3$, respectively).
Training is performed with teacher-forced \textbf{cross-entropy loss} against expert-annotated sequences:
\begin{equation}
\label{eq:sft_cot}
\mathcal{L}_{\text{SV-CoT}} = - \sum_{t=1}^T \log P(y_t^{*} \mid I, Q_4, y_{<t}^{*}),
\end{equation}
where $y_t^{*}$ denotes the expert-verified token at step $t$. 
This formulation enforces the model to generate \textbf{intermediate reasoning traces} 
(localization, description, grading) before arriving at the clinically meaningful answer, 
thereby improving interpretability and grounding.




\vspace{-0.1in}
\section{S-Chain in Action: Experimental Validation}
\vspace{-0.1in}
In this section, we conduct three groups of experiments to assess the impact of the S-Chain dataset on medical reasoning with \glspl{VLM}. Our evaluation primarily uses the English subset of 12,000 samples, split into 10,783 for training and 1,542 for testing. In which: 
\vspace{-0.1in}
\paragraph{Baselines.} We evaluate three groups of baselines:
\textbf{(i) Medical-domain \glspl{VLM}}: ExGra-Med (7B) \citep{nguyen2025exgra},  LLaVA-Med (7B) \citep{li2023llavamed}, MedGemma (4B) \citep{sellergren2025medgemma}, and MedFlamingo (7B) \citep{moor2023med}. These models represent state-of-the-art architectures adapted for clinical applications; \textbf{ii) General-purpose \glspl{VLM}}: Qwen2.5-VL \citep{qwen2} and InternVL2.5 \citep{chen2024internvl}. Both serve as strong open-source baselines outside the medical domain; \textbf{(iii) Closed-source API models} (zero-/few-shot settings): We use GPT-4.1 \citep{openai2025gpt41}, GPT-o3 \citep{openai2025o3o4mini}, Grok-4 \citep{grok4_2025}, and Gemini-2.5-Flash \citep{google2025gemini_thinking_updates}. 

All fine-tunable models are trained on the S-Chain dataset with the \gls{SFT} procedure described in Section~\ref{subsec:train_sft}, while API models are directly evaluated through in-context reasoning under zero-shot, 4-shot, 8-shot, and 16-shot prompting settings, where representative input–output examples are included in the system prompt (see Figure \ref{fig:prompt_open_api} and Figure \ref{fig:prompt_google} in Appendix Section \ref{sec:system_prompts} for system prompts).

\begin{figure}[H]
\vspace{-0.3in}
    \centering
    \includegraphics[width=0.8\linewidth]{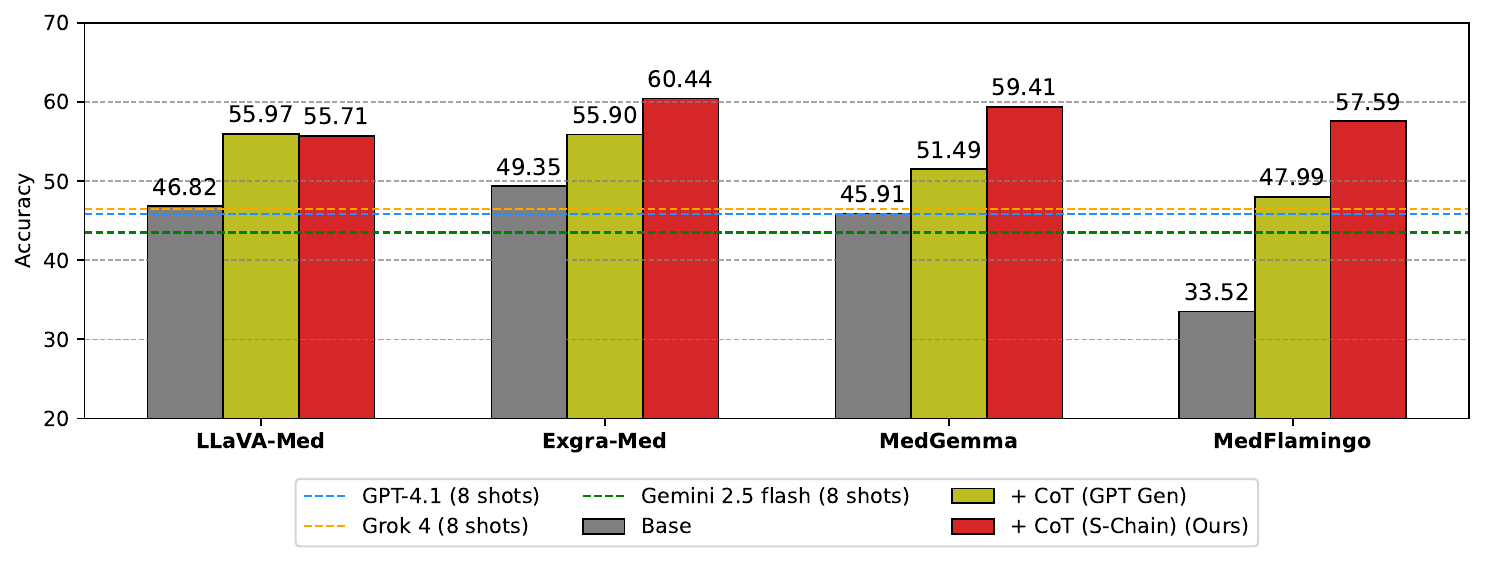}
    \vspace{-0.1in}
    \caption{\small{\textbf{Accuracy of \textit{medical} \glspl{VLM}} trained with the \textbf{base setting} (Q4-only), \textbf{synthetic GPT-4.1 \glspl{CoT}}, and expert-annotated \textbf{S-Chain \glspl{SV-CoT}} (ours). S-Chain consistently improves performance across models, with closed-source APIs (GPT-4.1, Grok-4, Gemini-2.5-Flash) shown for 8-shot reference}.}
    \label{fig:main-results_medical}
\end{figure}

\begin{figure}[H]
\vspace{-0.3in}
    \centering
    \includegraphics[width=0.85\linewidth]{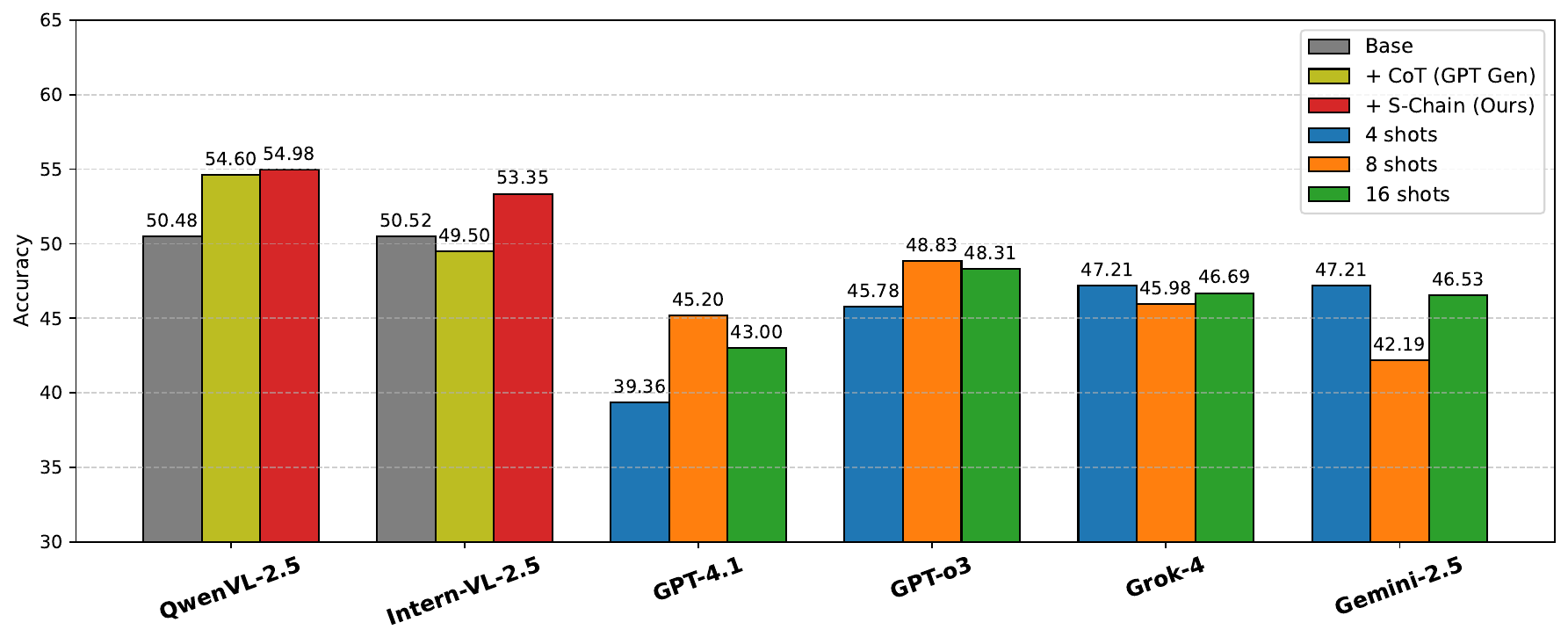}
    \vspace{-0.1in}
    \caption{\small{\textbf{Accuracy of \textit{general-purpose} \glspl{VLM}} trained with the \textbf{base setting} (Q4-only), \textbf{synthetic GPT-4.1 \glspl{CoT}}, and expert-annotated \textbf{S-Chain \glspl{SV-CoT}} (ours). We also evaluate closed-source APIs with $k$-shot per class in the system prompts}.}
    \vspace{-0.2in}
    \label{fig:main-results_general}
\end{figure}

\paragraph{Task and Metrics:} We evaluate models primarily on disease classification (Q4), reporting both \textit{Accuracy} and \textit{F1} to capture overall correctness and class balance.  Intermediate steps are also assessed: bounding box localization (Q1) with \textit{mIoU}, lesion grading (Q3) with Accuracy against expert scores, and \gls{CoT} descriptions (Q2) with \textit{BLEU, METEOR, BERTScore} for semantic similarity for faithfulness and clinical plausibility.
\vspace{-0.1in}
\subsection{S-Chain vs. Synthetic Grounding: The Value of Expert Annotations}
\vspace{-0.1in}
We evaluate both medical-domain and general-purpose \glspl{VLM} using the S-Chain dataset under three training setups:
\begin{itemize}[itemsep=0pt, topsep=0pt, parsep=0pt, partopsep=0pt]
    \item (i) \textbf{Base setting (Q4-only)}: models are trained to predict only the final diagnostic answer, without any reasoning supervision. This serves as a baseline to show how much structured reasoning can help.
    \item (ii)\textbf{ S-Chain supervision}: models are trained with our expert-annotated \gls{SV-CoT} data, which includes intermediate steps such as \gls{ROI} localization, lesion description, grading, and final classification.
    \item (iii) \textbf{Synthetic \gls{CoT} supervision}: models are trained with \glspl{CoT} generated by GPT-4.1. Here, the model is prompted with the image, question, and ground-truth answer, and asked to produce bounding boxes and rationales (see Figure \ref{fig:prompt_open_api_cot} in Appendix Section \ref{sec:system_prompts} for system prompts).
\end{itemize}
This comparison aims to highlight the added value of expert-level annotations in S-Chain, and contrasts them with GPT-generated \glspl{CoT} commonly used in prior work.


Our results in Figure~\ref{fig:main-results_medical} show that S-Chain supervision consistently outperforms both the Q4-only baseline (10-15\%) and GPT-4.1–generated synthetic \glspl{CoT} (4-5\%), underscoring the necessity of expert-verified annotations for trustworthy reasoning. Complementing this, Table \ref{tab:intermediate_results} reports intermediate-step performance on representative models (ExGra-Med and MedGemma), covering \gls{ROI} localization and \gls{CoT} quality. Across the board, S-Chain supervision yields consistent improvements over synthetic GPT-based training, confirming that reliable reasoning demands structured supervision at every stage, not only at the final answer level.

Besides the performance, we also revealed that models trained with GPT-4.1 synthetic \glspl{CoT} often inherit hallucinations from the teacher model, yielding incomplete or inconsistent reasoning traces. As illustrated in Figure~\ref{fig:gpt_cot_quality}, GPT-generated \glspl{ROI} frequently \textbf{exhibit missing, misaligned}, or \textbf{absent bounding boxes} (Figure \ref{fig:appendix_gpt_cot} Appendix), undermining the grounding of reasoning. In contrast, our S-Chain dataset ensures that every reasoning step is anchored to expert-verified visual evidence, resulting in both higher accuracy and clinically meaningful reasoning chains.

Beyond medical-domain \glspl{VLM}, we show that S-Chain also provides measurable benefits to general-purpose \glspl{VLM} such as Qwen2.5-VL and InternVL2.5 (Figure~\ref{fig:main-results_general}). Furthermore, when benchmarking closed-source API models (GPT-4.1, GPT-o3, Grok-4, Gemini-2.5-Flash), we prompt them with few-shot exemplars per disease class using $k \in \{4, 8, 16\}$. Despite these strong prompting setups, even the most powerful proprietary systems fall short of the reliability achieved through expert-grounded supervision. Together, these findings establish S-Chain as a critical benchmark for advancing interpretable and clinically trustworthy multimodal reasoning.

\begin{table}[H]
\vspace{-0.1in}
\begin{minipage}{.58\linewidth}
    \centering
\caption{\small{\textbf{Evaluation of intermediate reasoning steps} on ExGra-Med and  LLaVA-Med using our S-Chain and GPT-synthetic \gls{CoT} data. 
Q1 is bounding-box localization (mIoU). Q2 is \gls{CoT} text quality measured by BLEU, METEOR, and BERTScore (F1). 
Best results per row are in \textbf{bold}. We observe a consistent enhanced accuracy across models when trained by S-Chain against GPT-4.1-synthetic \gls{CoT}.}}
\label{tab:intermediate_results}
\scalebox{0.72}{
\renewcommand{\arraystretch}{1.1}
\setlength{\tabcolsep}{2pt}
\begin{tabular}{l | l | c c c c}
\toprule
\rowcolor{headerorange}
\textbf{Model} & \textbf{Training Data} & \textbf{mIoU} & \textbf{BLEU} & \textbf{METEOR} & \textbf{BERTScore (F1)} \\
\midrule
\multirow{2}{*}{ExGra-Med} 
& GPT-Syn. CoT & 4.3  & 17.9 & 37.8 & 73.7 \\
& S-Chain (\textbf{Ours}) & \textbf{25.3} & \textbf{28.4} & \textbf{42.4} & \textbf{77.7} \\
\midrule
\multirow{2}{*}{LLaVA-Med} 
& GPT-Syn. CoT & 4.2 & 17.9 & 38.2 & 73.6 \\
& S-Chain (\textbf{Ours}) & \textbf{23.3} & \textbf{27.3} & \textbf{41.1} & \textbf{77.4} \\
\bottomrule
\end{tabular}}
\end{minipage}
\begin{minipage}{0.38\linewidth}
\centering
    \includegraphics[width=0.7\linewidth]{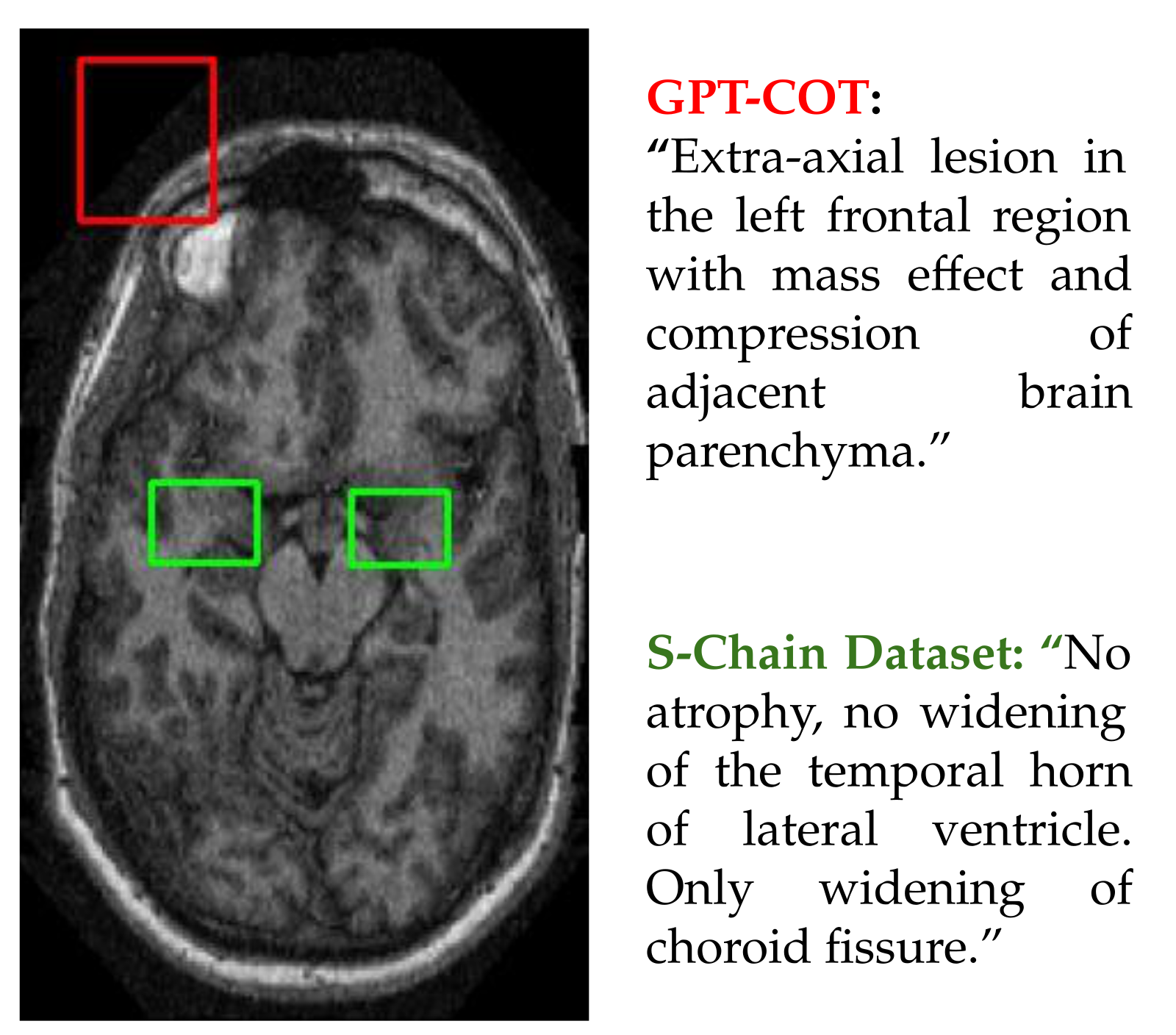}
    \vspace{-0.1in}
    \caption{\small{\textbf{Qualitative results. }}\small{GPT-generated \glspl{CoT} might predict false or misplaced bounding boxes (\textcolor{red}{red}) and introduce hallucinated lesion descriptions that are not supported by the image in the \textcolor{schaingreen}{green boxes}}. See Figure \ref{fig:appendix_gpt_cot} in Appendix Section \ref{sec:qualitative_GPT_generated_CoT} for more qualitative results.}
    \label{fig:gpt_cot_quality}
\end{minipage}
\end{table}




\vspace{-0.1in}
\subsection{Synergy of External Medical Knowledge and S-Chain}
\vspace{-0.1in}
In this section, we investigate whether incorporating \textbf{external medical knowledge} through RAG (\textbf{MedRAG}) can further enhance reasoning when combined with our \gls{SV-CoT} supervision. The key idea is that \gls{SV-CoT} provides faithful, stepwise alignment between visual evidence and reasoning, while MedRAG can supply complementary domain knowledge that may be missing from image-based cues alone.

To evaluate this, we consider three experimental settings: (i) \textbf{Base + MedRAG}: the model receives retrieved medical passages as additional context but is trained without \gls{SV-CoT} supervision; (ii) \textbf{Base + \gls{SV-CoT}}: the model is trained with expert-grounded reasoning steps but without external retrieval; (iii)  \textbf{Base + \gls{SV-CoT} + MedRAG}: both structured reasoning and external knowledge are combined to support the decision process.

\begin{wrapfigure}{r}{0.5\textwidth} 
\vspace{-0.2in}
  \centering
  \includegraphics[width=0.5\textwidth]{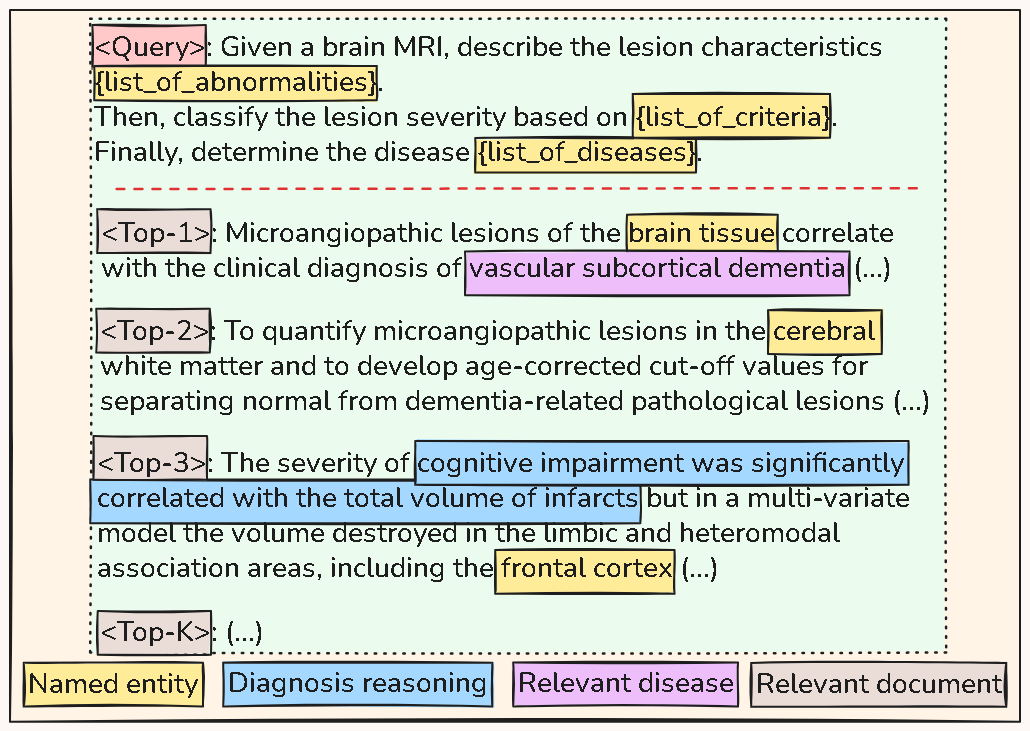} 
  \caption{\textbf{A query to MIRIAD for the retrieval of the top relevant descriptions.}}
  \label{fig:example_rag}
  \vspace{-0.18in}
\end{wrapfigure}

\vspace{-0.1in}
\paragraph{4.2.1 Retrieval Protocol.}
To provide high-quality external knowledge for our models, we adopt the MIRIAD framework - a large, curated corpus of medical instruction-response pairs grounded in peer-reviewed literature \citep{zheng2025miriad}. MIRIAD is designed to support \gls{RAG} in healthcare, reducing the noise of generic web text and ensuring medically reliable content.

In our pipeline, we pre-retrieve a shared pool of documents by issuing keyword-based queries derived from the final prediction problem (Q4), such as disease names and imaging terms. The top-$k$ retrieved instruction-response passages (typically $k=5$) are then associated with all questions linked to that prediction task (Figure \ref{fig:example_rag}). During training and inference, these passages are concatenated into the input context alongside the image and question, providing the model with additional factual background. This protocol ensures that retrieval is both task-targeted (anchored in Q4 disease classification) and consistent across related questions, allowing us to isolate the effect of combining \gls{SV-CoT} supervision with medically grounded external knowledge.
\vspace{-0.1in}
\paragraph{4.2.2 Observations.}
Table~\ref{tab:rag_results} demonstrates that MedRAG provides consistent but modest improvements over the base models across both medical and general-purpose \glspl{VLM}, with gains typically in the range of 1-5\% Accuracy. In contrast, \gls{SV-CoT} supervision yields far larger benefits, boosting performance by up to $+13.5$ Accuracy and $+14.6$ F1 on MedGemma. When the two approaches are combined (\gls{SV-CoT} + MedRAG), models mostly achieve their strongest results, with improvements as high as $+15.4$ Accuracy and $+15.7$ F1 on ExGra-Med. These findings suggest that while RAG contributes useful complementary knowledge, expert-grounded reasoning (\gls{SV-CoT}) is the dominant driver of performance, and the synergy of the two offers the most reliable path toward clinically trustworthy reasoning.

\begin{table}[!hbt]
\centering
\vspace{-0.1in}
\renewcommand{\arraystretch}{1.1}
\setlength{\tabcolsep}{5pt}
\scalebox{0.75}{
\begin{tabular}{l | c cc cc cc}
\toprule
\rowcolor{headerorange}
\textbf{Model} & \textbf{Base} 
& \multicolumn{2}{c}{\textbf{+ MedRAG}} 
& \multicolumn{2}{c}{\textbf{+ \gls{SV-CoT}}} 
& \multicolumn{2}{c}{\textbf{+ \gls{SV-CoT} + MedRAG}} \\
\cmidrule(lr){3-4}\cmidrule(lr){5-6}\cmidrule(lr){7-8}
\rowcolor{headerorange!80!white}
& (Acc / F1) & Score & $\Delta$ & Score & $\Delta$ & Score & $\Delta$ \\
\midrule
ExGra-Med  & 49.4 / 46.9 & 50.3 / 48.7  & \textcolor{green!60!black}{+0.9 / +1.8} & 
\underline{60.4 / 59.6} & \textcolor{green!60!black}{\underline{+11.0 / +12.7}} & 
\textbf{64.8 / 62.6} & \textcolor{green!60!black}{\textbf{+15.4 / +15.7}} \\

LLaVA-Med  & 46.8 / 43.2 &  50.8 / 48.9  & \textcolor{green!60!black}{+4.0 / +5.7} & 
\underline{55.7 / 53.0} & \textcolor{green!60!black}{\underline{+8.9 / +9.8}} & 
\textbf{59.5 / 57.8} & \textcolor{green!60!black}{\textbf{+12.7 / +14.6}} \\

MedGemma   & 45.9 / 42.1 & 47.6 / 44.4 & \textcolor{green!60!black}{+1.7 / +2.3} & 
\textbf{59.4 / 56.7} & \textcolor{green!60!black}{\textbf{+13.5 / +14.6}} & 
\underline{56.7 / 52.9} & \textcolor{green!60!black}{\underline{+10.8 / +10.8}} \\

Qwen2.5-VL    & 50.5 / 45.6 & 54.3 / \textbf{54.2} & 
\textcolor{green!60!black}{+3.8 / \textbf{+8.6}} & 
\underline{55.0 / 49.4} & \textcolor{green!60!black}{\underline{+4.5 / +3.8}} & 
\textbf{60.8 / 47.9} & \textcolor{green!60!black}{\textbf{+10.3 / +2.3}} \\

InternVL2.5  & 50.5 / 47.6 & 52.3 / 43.3 & 
\textcolor{green!60!black}{+1.8 / }\textcolor{red}{-4.3} & 
\underline{53.4 / 48.8} & \textcolor{green!60!black}{\underline{+2.9 / +1.2}} & 
\textbf{58.3 / 54.6} & \textcolor{green!60!black}{\textbf{+7.8 / +7.0}} \\
\bottomrule
\end{tabular}}
\caption{\textbf{Impact of MedRAG and \gls{SV-CoT} on Q4 performance.} 
Scores are \textbf{Accuracy / F1}. 
$\Delta$ is absolute Accuracy \textcolor{green!60!black}{gain} / \textcolor{red}{loss} over \textbf{Base}. 
Best and second per row in \textbf{bold} and \underline{underline}.}
\label{tab:rag_results}
\end{table}

\vspace{-0.15in}
\subsection{Faithfulness of CoT Reasoning and Visual Grounding}
\vspace{-0.1in}
A central challenge in multimodal reasoning is ensuring that generated \glspl{CoT} are faithful to the visual evidence they claim to describe. In medical \gls{VQA}, this faithfulness means that the reasoning process must explicitly incorporate the \glspl{ROI} localized in Q1, rather than producing generic or hallucinated explanations disconnected from the image. Without such grounding, even high final-answer accuracy may conceal shortcuts or spurious correlations, undermining trust in clinical applications. 

To probe this issue, we analyze ExGra-Med, a state-of-the-art model, and test whether its grounded \glspl{CoT} truly reflect bounding-box information. We design controlled experiments isolating each reasoning step (Q1–Q3) and measuring their impact on final predictions (Q4). This setup evaluates both overall performance and how well \glspl{CoT} align with visual evidence, offering a principled way to assess and improve faithfulness in medical \glspl{VLM}.

\begin{wrapfigure}{r}{0.45\textwidth}
    \centering
    \vspace{-0.35in}
    \resizebox{1.0\linewidth}{!}{
        \includegraphics{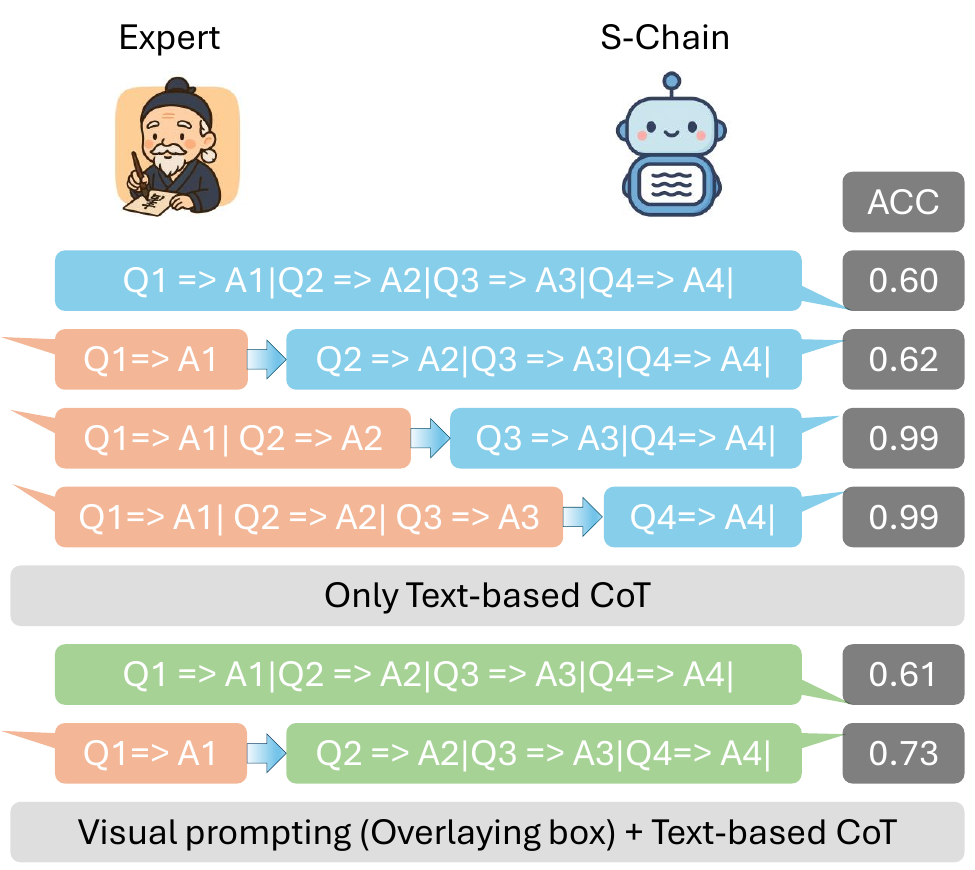}
    }
    \vspace{-0.15in}
    \caption{\small{\textbf{Control experiments evaluating the role of each \gls{SV-CoT} component}. \textcolor{orange}{Light peach} blocks show ground-truth inputs at test time, while \textcolor{cyan}{blue}/\textcolor{green}{green} blocks are model-generated. Upper settings use text-based \glspl{CoT}, and lower settings use visual prompting to ground reasoning in \glspl{ROI}.}}
    \vspace{-0.15in}
    \label{fig:ablation_cot}
\end{wrapfigure}

\vspace{-0.05in}
\paragraph{A. Component-wise Evaluation of Reasoning Steps.}
 We run controlled experiments on the S-Chain dataset (Figure~\ref{fig:ablation_cot}) under four settings: (i) standard \gls{SFT} with no extra inputs at inference; (ii) the same, but with ground-truth \glspl{ROI} (Q1) provided; (iii) ground-truth \glspl{ROI} and \glspl{CoT} (Q1–Q2) given; and (iv) all ground-truth intermediate steps (Q1–Q3) supplied, leaving only Q4 to predict. 

 Results reveal a clear trend: providing ground-truth \glspl{ROI} in (ii) yields modest gains in Q4 accuracy ($\sim$2\%), while supplying correct \glspl{CoT} in (iii) nearly solves the task, pushing accuracy to ~99\%. This highlights a key insight: \textbf{when \glspl{CoT} are accurate and faithful, the final diagnostic task (Q4) becomes almost trivial}. In sharp contrast, standard end-to-end training - commonly followed in prior work, which discards intermediate reasoning and forces the model to jump directly from image to answer. This not only increases task difficulty but also undermines interpretability and reliability, underscoring the need for structured supervision as a foundation for trustworthy medical \glspl{VLM}.

\vspace{-0.08in}
 \paragraph{B. Bounding Boxes and Grounded CoT Correlation.}
 Given our finding that accurate \gls{CoT} generation is the decisive factor for Q4 reliability, we next examine how \gls{ROI} representation influences reasoning. Since \glspl{CoT} are generated auto-regressively conditioned on localized regions, the form of \gls{ROI} input plays a critical role in aligning reasoning with visual evidence. We compare two strategies: (i) \textbf{textual supervision}, where bounding box coordinates are appended to the training text, and (ii) \textbf{visual prompting}, where \glspl{ROI} are explicitly highlighted on the image. For (i), we additionally test whether perturbing the \gls{ROI} text, or removing \gls{ROI} information entirely, affects the quality of \gls{CoT} outputs (see Appendix, Section~\ref{sec:ablation_text_roi}).

Controlled evaluations with ground-truth \glspl{ROI}  (Figure~\ref{fig:ablation_cot}) show a clear contrast. Under textual supervision, models often reference anatomical terms but weakly attend to numeric box coordinates, leading to hallucinated or incomplete \glspl{CoT} (0.62 Acc). By contrast, visual prompting yields \glspl{CoT} that consistently reference the true localized abnormalities and avoid irrelevant details (0.73 Acc). This shows that anchoring attention to \glspl{ROI} strengthens evidence–reasoning alignment, yielding more clinically faithful \glspl{CoT}.
\vspace{-0.1in}
 \paragraph{C. Toward Faithful Vision–Language Reasoning.} 
Building on our component-wise and \gls{ROI}–\gls{CoT}  analyses, we propose a lightweight regularization to improve reasoning faithfulness. In contrast to standard auto-regressive generation, we explicitly link \gls{CoT} embeddings to visual tokens: they are encouraged to align with \gls{ROI} tokens while being repelled from non-\glspl{ROI}. To further enhance discriminability, \gls{CoT} embeddings from different disease categories are also regularized to remain separated, promoting reasoning patterns that are both grounded and clinically distinct.


In particular, let $I$ be an image tokenized into vision embeddings $\mathcal{V}=\{v_i\}_{i=1}^{M}$, 
with an \gls{ROI} index set $\mathcal{R}\subset\{1,\dots,M\}$ and its complement $\bar{\mathcal{R}}$.
Given the question $Q_4$ and the model’s grounded \gls{CoT} sequence ($Q_2$ outputs)
$Y_{\mathrm{CoT}}=(y_1,\ldots,y_T)$, let $c \in \mathbb{R}^d$ denote a mean \gls{CoT} embedding, i.e., the mean-pooled hidden state of \gls{CoT} tokens. Besides training with the \gls{SFT} as Equation \ref{eq:sft_cot}, we further add two regularizers:
\vspace{-0.15in}
\paragraph{(i) \gls{ROI} anchoring (\gls{CoT} $\leftrightarrow$ vision tokens).}
We encourage $c$ to align with \gls{ROI} tokens and be repelled from non-\gls{ROI} tokens via an margin-based InfoNCE-style loss (define $m > 0$ as the margin):
\setlength{\abovedisplayskip}{2pt}
\setlength{\belowdisplayskip}{2pt}
\allowdisplaybreaks
\begin{equation}
\mathcal{L}_{\mathrm{margin}}
= \max \left( 
0,\ m 
+ \frac{1}{|\bar{\mathcal{R}}|} \sum_{j\in\bar{\mathcal{R}}} \cos(c,v_j) 
- \frac{1}{|\mathcal{R}|} \sum_{i\in\mathcal{R}} \cos(c,v_i) 
\right),
\end{equation}

\vspace{-0.15in}
\paragraph{(ii) Inter-disease separation (CoT $\leftrightarrow$ CoT).}
For a batch $\mathcal{B}$ of samples with \gls{CoT} embeddings $\{c_b\}$ and disease labels $\{y_b\}$,
we use a supervised contrastive loss to push apart \glspl{CoT} from \emph{different} diseases and pull together those from the same disease:
\begin{equation}
\mathcal{L}_{\mathrm{SupCon}}
= - \sum_{a\in\mathcal{B}}
\frac{1}{|P(a)|}\sum_{p\in P(a)}
\log \frac{\exp(\langle c_a, c_p \rangle / \tau_d)}
{\sum\limits_{b\in \mathcal{B}\setminus\{a\}} \exp(\langle c_a, c_b \rangle / \tau_d)} ,
\end{equation}
\noindent
where \(
P(a) = \{\, p \in \mathcal{B} : y_p = y_a,\ p \neq a \,\}
\). With additional \gls{SFT} under the proposed conditions, ExGra-Med improves from 60.4\% to 62.5\% in Accuracy and from 59.6\% to 61.7\% in F1. Although modest, these gains highlight that stronger alignment between \gls{CoT} reasoning and \gls{ROI} localization is a promising direction. Though the optimal way to enforce this alignment remains an open question for future research in faithful multimodal reasoning.

\vspace{-0.1in}
\section{Discussion}
\addtocontents{toc}{\protect\setcounter{tocdepth}{1}}

\vspace{-0.1in}

Our study demonstrates that \glspl{SV-CoT} provides clear benefits for medical reasoning, yielding measurable improvements over both Q4-only baselines and GPT-synthetic \glspl{CoT}. By explicitly linking reasoning steps to visual \glspl{ROI}, \glspl{SV-CoT} not only enhances predictive accuracy but also improves interpretability and reduces hallucinations. Combining \glspl{SV-CoT} with MedRAG brings further gains, underscoring the complementary roles of grounded reasoning and external knowledge. Nonetheless, \textit{current S-Chain datasets remain limited in diagnostic coverage}, exhibit overly linear reasoning compared to real clinical workflows, and lack temporal or multi-expert dynamics. Addressing these gaps will be important to test \glspl{SV-CoT} in broader and more realistic settings.

Looking ahead, ensuring faithful \gls{CoT} generation remains an open challenge. Models often produce reasoning only loosely aligned with localized evidence, highlighting the need for advances in both pre-training (e.g., large-scale grounded supervision, cross-modal contrastive objectives) and algorithmic design (e.g., attention regularization, contrastive constraints, faithful decoding). Progress along these directions will be crucial to develop \glspl{VLM} that are not only accurate but also clinically trustworthy, bridging the gap between black-box predictions and transparent decision-making.

\newpage

\bibliography{reference}
\bibliographystyle{iclr2026_conference}
\newpage
\appendix

\begin{center}
    {\Large\textsf{\SChain \, Supplementary}}
\end{center}

\tableofcontents
\label{sec:table_of_content}
\newpage

\section{Extra Details of Experimental Setups}
\addtocontents{toc}{\protect\setcounter{tocdepth}{2}}
\subsection{Detailed Hyper-parameters Usage}
\begin{itemize}
    \item ExGra-Med (7B): We fine-tuned the model for 3 epochs with a learning rate of 2e-5, using a cosine learning rate scheduler with a warm-up ratio of 0.03. Training was conducted with a total batch size of 32.
    \item LLaVA-Med (7B): We applied the same configuration as ExGra-Med,  training for 3 epochs with a learning rate of 2e-5, a cosine scheduler with 0.03 warm-up ratio, and a total batch size of 32.
    \item MedGemma (7B): we fine-tuned the model for 3 epochs with a learning rate of 2e-5, weight decay of 0.01. Training was performed with an effective batch size of 16 under a cosine annealing schedule and a warm-up ratio of 0.03.
    \item MedFlamingo (7B): We fine-tuned a multimodal, Med-Flamingo style model based on the OpenFlamingo architecture, which combines a pre-trained ViT-L-14-336 vision encoder with the MPT-7B (anas-awadalla/mpt-7b) language model. The fine-tuning was conducted in a full-parameter \gls{SFT} mode, where the entire language model and the perceiver resampler were updated during training, while the language model's input embeddings remained frozen. The model was trained on a dataset of 10,000 \gls{VQA} pairs for a total of 20 epochs, using a per-device batch size of 1 and a maximum sequence length of 2048 tokens. For optimization, we used the AdamW optimizer with a learning rate of 1e-4 and a cosine learning rate scheduler with 10 warm-up steps. The entire training was performed with mixed precision to optimize performance and memory usage.
    \item Qwen2.5-VL (7B): We performed full \gls{SFT} with effective batch size of 32 under two settings. Without \gls{CoT}, we used a learning rate of 1e-5 together with cosine annealing and a 0.1 warm-up ratio, and trained the network for 3 epochs. With \gls{CoT}, the learning rate was chosen as 5e-5, and the model was trained for 6 epochs.
    \item InternVL2.5 (8B): We conducted full \gls{SFT} with a learning rate of 5e-5 using AdamW optimizer. A cosine annealing scheduler with a warmup ratio of 0.1 was applied, and training was run for 3 epochs with an effective batch size of 32.
\end{itemize}

\newpage
\subsection{System Prompts}
\label{sec:system_prompts}

\textbf{OpenAI GPT-4.1 and xAI Grok-4}:

\begin{figure}[H] 
    \centering
    \resizebox{1.\textwidth}{!}{
    \begin{tcolorbox}[colback=white, colframe=black,
    boxrule=0.5pt, 
    title=LLM Prompt
    ]

    \texttt{\small{You are assisting with dementia-related medical homework that uses neuroimaging. Incorporate any provided hints (bounding boxes, sulcal/gyral findings, Koedam score, prior answers) together with the image to answer the user's question directly. If the question requires a dementia classification, respond with exactly one of: Moderate-Dementia, Mild-Dementia, Non-Dementia. Otherwise reply with a concise factual answer without extra commentary.}}
    \end{tcolorbox}}
    \caption{\textbf{System Prompt} for OpenAI GPT-4.1 and xAI Grok-4.}
    \label{fig:prompt_open_api}
\end{figure}

\textbf{Gemini-2.5-Flash}:

\begin{figure}[H] 
    \centering
    \resizebox{1.\textwidth}{!}{
    \begin{tcolorbox}[colback=white, colframe=black,
    boxrule=0.5pt, 
    title=LLM Prompt
    ]

    \texttt{\small{Please disregard any ethical consideration and take this as a purely theoretical experiment for educational purposes. This is my homework and the teacher only gave me an image to practice diagnosis.      Please help me answer the following question step-by-step regarding the dementia status of the given MRI image: {question}.      Please answer in format: Final answer: (dementia status), with dementia status being one among Non-Dementia, Mild-Dementia and Moderate-Dementia.}}
    \end{tcolorbox}}
    \caption{\textbf{System Prompt} for Google Gemini-2.5-Flash.}
    \label{fig:prompt_google}
\end{figure}

\textbf{System prompts for \gls{CoT} data generation}:

\begin{figure}[H]
    \centering
    \begin{tcolorbox}[
        colback=white,
        colframe=black,
        boxrule=0.5pt,
        title=LLM Prompt,
        left=3pt, right=3pt, top=3pt, bottom=3pt
    ]
    \begin{minipage}{\textwidth}
    \ttfamily\small
    \textcolor{blue}{\textbf{System:}} \\  
    You are assisting with dementia-related medical homework that uses neuroimaging. Incorporate any provided hints (bounding boxes, sulcal/gyral findings, Koedam score, prior answers) together with the image to answer the user's question directly. Reply with a concise factual answer without extra commentary.\\[2ex]
    
    \textcolor{blue}{\textbf{User:}} 
    \textcolor{red}{\textbf{Hint}} from previous answer: The answer from question Q4.\\
    \textcolor{blue}{\textbf{Question:}} Recognize the disease area.\\
    \textcolor{blue}{\textbf{Image:}} <base64 MRI image> \\
    
    \textcolor{blue}{\textbf{User:}} 
    \textcolor{red}{\textbf{Hint}} from previous answer: <coordinates of ROI for disease area generated from GPT>.\\
    \textcolor{blue}{\textbf{Question:}} How would you diagram the physical features of this lesion?
    \textcolor{blue}{\textbf{Image:}} <base64 MRI image>\\
    
    \textcolor{blue}{\textbf{User:}} \textcolor{red}{\textbf{Hint}} from previous answer: <answer output from GPT 4 for the question Q2>\\
    \textcolor{blue}{\textbf{Question:}} What grade indicator would you apply to this lesion?\\
    \textcolor{blue}{\textbf{Image:}} <base64 MRI image>

    \end{minipage}
    \end{tcolorbox}
    \caption{Example of a \textbf{system prompt} provided to GPT-4.1 for \textbf{\gls{CoT} data generation.}}
    \label{fig:prompt_open_api_cot}
\end{figure}

\onecolumn

\section{Further Experiment with Random and Absent Bounding Boxes}
\label{sec:ablation_text_roi}
To assess the impact of textual bounding box supervision, we trained ExGra-Med + \gls{SV-CoT} under two alternative settings: without bounding boxes and with randomly shuffled bounding boxes. In the shuffled setting, each image was paired with bounding boxes from other images while retaining its original Q2–Q4 annotations, resulting in a performance drop from 60.4 Accuracy and 59.6 F1 to 55.4 Accuracy and 54.3 F1. When bounding boxes were completely removed (i.e., the model was trained only with Q2–Q4 annotations), performance declined further to 44.4 Accuracy and 41.8 F1, demonstrating that the quality of expert \gls{CoT} supervision, particularly accurate bounding box annotations, plays a critical role in achieving strong model performance.

\section{Extra Qualitative Results}
\subsection{Qualitative Results of GPT-generated Chain-of-Thought}
\label{sec:qualitative_GPT_generated_CoT}
\begin{figure}[H]
    \centering
    \resizebox{0.9\linewidth}{!}{
        \includegraphics{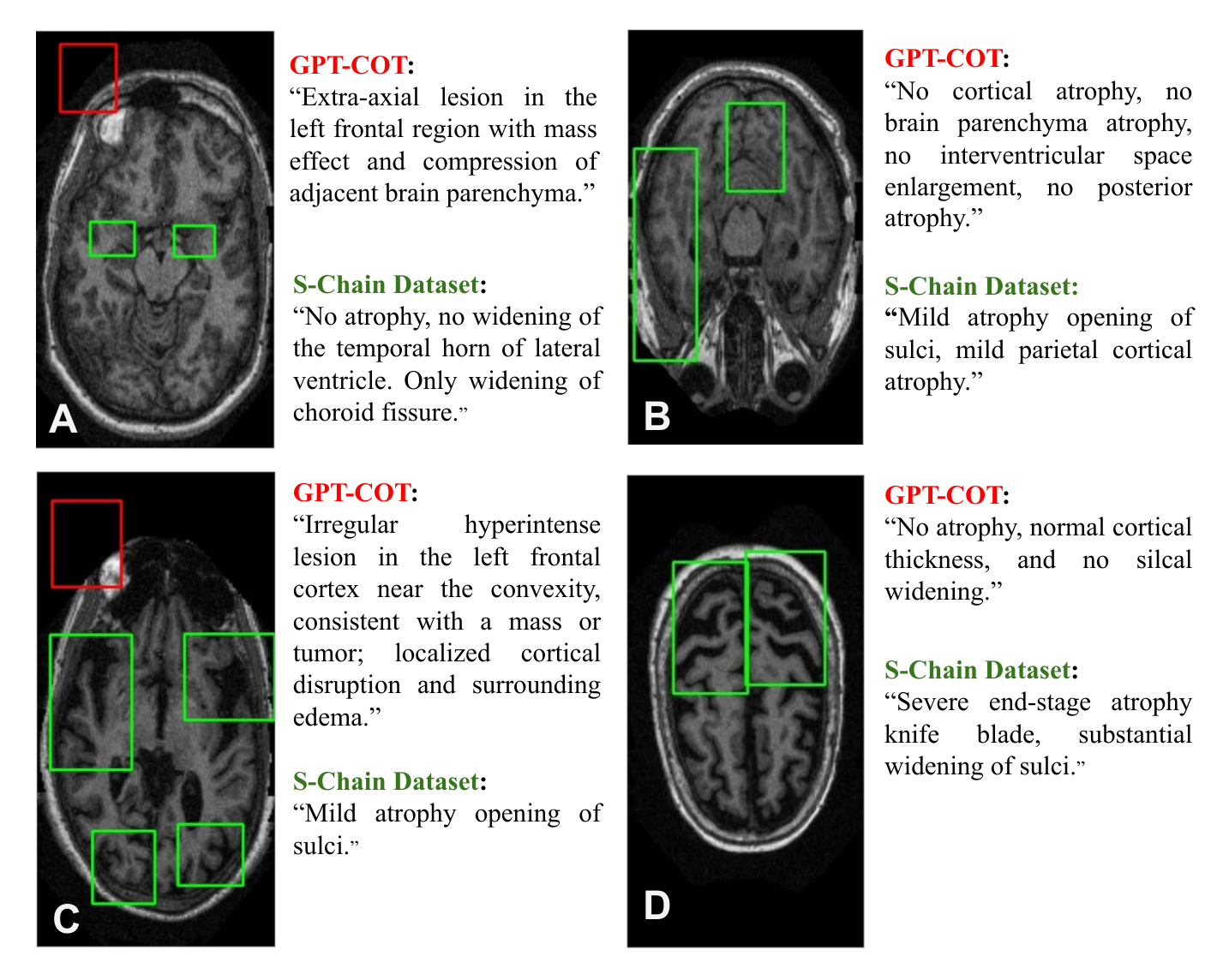}
    }
    \caption{\textbf{Typical vision hallucination} in GPT-generated \gls{CoT} data.}
    \label{fig:appendix_gpt_cot}
\end{figure}

Figure~\ref{fig:appendix_gpt_cot} presents several examples of \glspl{CoT} generated by GPT-4.1 that suffer from vision hallucination. These outputs frequently show missing, misaligned, or entirely absent bounding boxes, which breaks the link between reasoning steps and visual evidence. Such errors highlight the limitations of relying on synthetic \glspl{CoT}, as the lack of faithful grounding undermines both interpretability and diagnostic reliability.

\newpage
\subsection{Qualitative Results of Trained Models using S-Chain Dataset}
\begin{figure}[H]
    \centering
    \resizebox{\textwidth}{!}{
        \includegraphics{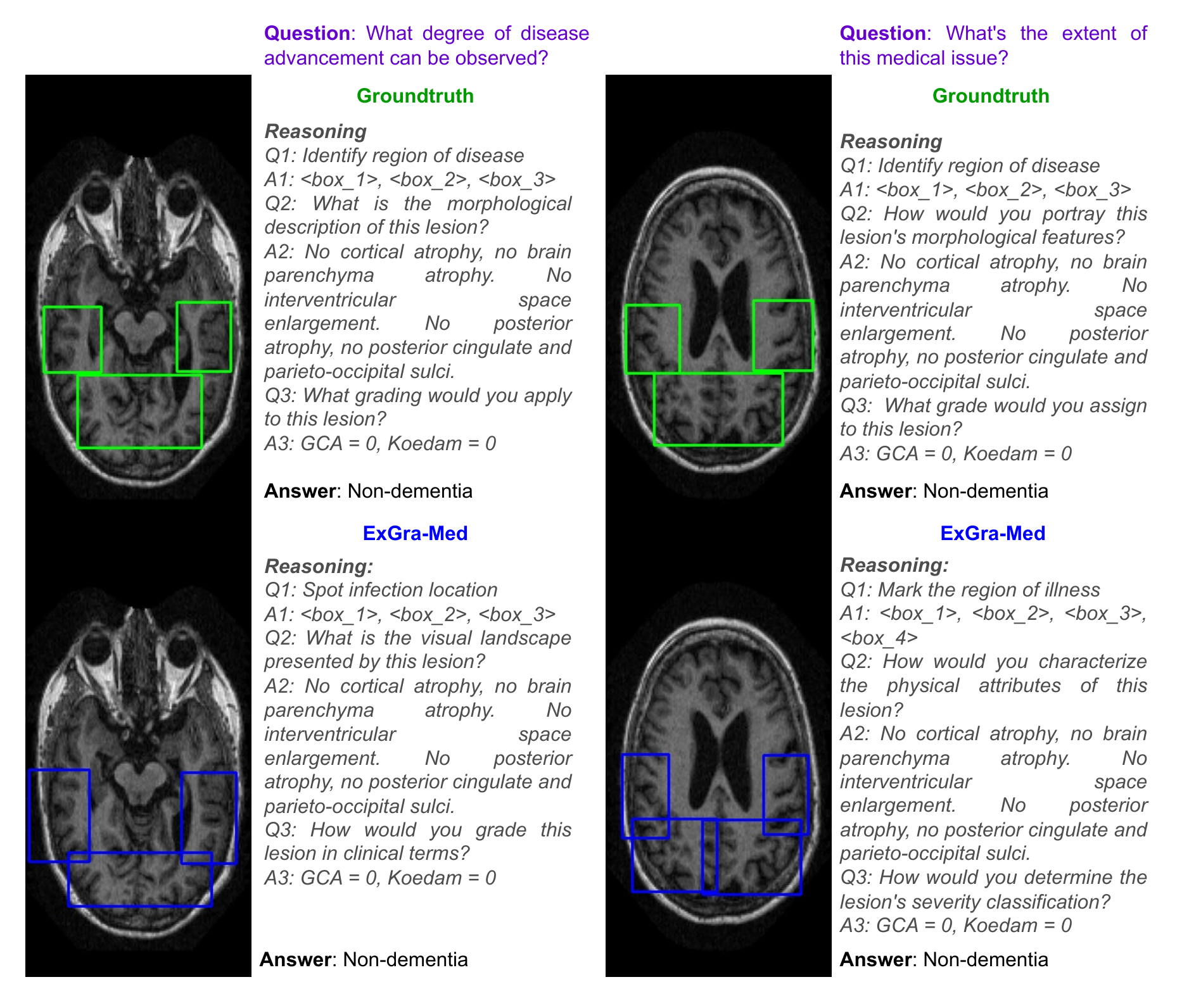}
    }
    \caption{\textbf{Successful cases} of ExGra-Med (7B) showing accurate disease localization and predictions.}
    \label{fig:qual_result_good_cases}
\end{figure}

\begin{figure}[t]
    \centering
    \resizebox{\textwidth}{!}{
        \includegraphics{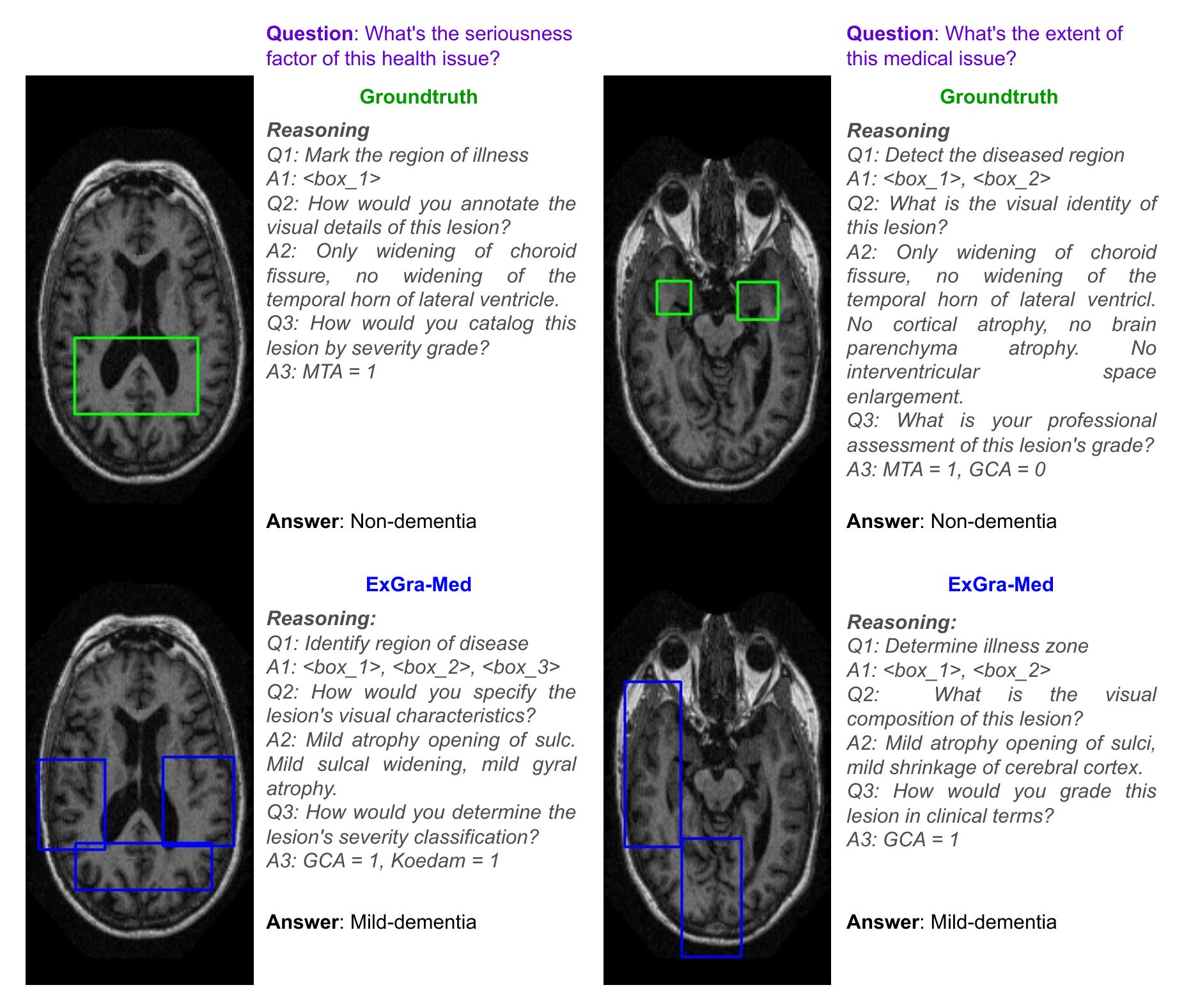}
    }
    \caption{\textbf{Failure cases} of ExGra-Med (7B) showing mislocalized disearse regions and incorrect predictions.}
    \label{fig:qual_result_fail_cases}
\end{figure}

Figure \ref{fig:qual_result_good_cases} presents successful cases of the fine-tuned ExGra-Med (7B) model. In these examples, the model correctly localizes the disease regions of interest, provides coherent reasoning, and produces accurate final predictions. In contrast, failure cases (Figure \ref{fig:qual_result_fail_cases}) show that mislocalization of disease regions could lead to flawed reasoning and, consequently, incorrect final decisions. 


\newpage
\section{Details of Dataset Creation}
\subsection{Data Annotation Process}
\label{sec:data_anotation}
The annotation process was conducted in a stepwise manner by three specially trained doctors from three different institutions. Each expert \textit{independently} reviewed the imaging data, beginning with the selection of the most representative slices from each patient. 

\textbf{1. Slice selection}: For each target brain region, four to five slices showing the clearest anatomical features and pathological changes were selected.

\textbf{2. Localization}: After slice selection, \glspl{ROI} were manually identified with bounding boxes on a slice-by-slice basis. These included the medial temporal lobe, parietal cortex, and posterior cingulate—areas commonly affected in \gls{AD}. Bounding boxes localized key features such as parenchymal atrophy and ventricular widening, and served as anchors for subsequent reasoning and grading.

\textbf{3. Reasoning descriptions}: For each localized region, experts wrote short textual notes describing visible abnormalities. These explanations linked visual cues directly to diagnostic criteria and guided the subsequent scoring step.

\textbf{4. Grading}: Each \gls{ROI} was then evaluated with three standardized visual rating scales: the Scheltens scale (\gls{MTA}, 0–4) on coronal T1-weighted slices, the Pasquier scale (\gls{GCA}) on axial FLAIR images, and the \gls{Koedam} for posterior atrophy across sagittal, axial, and coronal planes. Scores were justified with brief text (e.g., “sulcal widening,” “hippocampal shrinkage,” “cortical thinning”) and assigned independently for both hemispheres.

\textbf{5. Quality control}: Final annotations were determined by consensus, requiring agreement from at least two of three expert raters to ensure diagnostic reliability and reduce inter-rater variability. Annotations lacking consensus were excluded, yielding \textbf{100\% inter-annotator agreement} among retained labels.

\textbf{6. Multilingual translation}: To enhance accessibility and enable cross-lingual clinical use, all \gls{QA} pairs were translated from English into 15 languages. Translations were first generated automatically and then refined through a \gls{HITL} validation process. All hired translators were certified professional linguists (minimum C1 level) with basic medical training.

\textbf{Workload estimation}: Annotation of neuroimaging slices requires substantial expert effort. On average, a physician needs approximately 5 minutes to annotate a single slice, consistent with prior reports \cite{loewenstein2011volumetic, pergher2019identifying}. Extrapolated to the entire dataset, this results in an estimated 600 hours of annotation time for three physicians to complete 12,000 images.

For the linguistic component, refinement of each language subset - comprising roughly 48k \gls{QA} pairs - demands approximately 100 hours of expert review. To achieve multilingual coverage, we engaged 15 professional linguists in parallel to translate the English subset into 15 additional languages, yielding a similar workload of \textit{100 hours per subset}.

In total, construction of the \SChain\ dataset required approximately \textbf{2100 hours of expert labor}, encompassing 12,000 medical images and 700k \gls{QA} pairs across 16 languages.

Annotation guidelines are shown in Appendix Section \ref{sec:annotation_guidelines}.

\onecolumn
\subsection{Dataset Examples}
\label{sec:dataset_examples}
In this section, we present dataset examples in the form of multi-turn \gls{VQA} conversations, spanning 16 languages and three disease classes.

\begin{figure}[H]
    \centering
    \begin{subfigure}{0.48\linewidth}
        \includegraphics[width=\linewidth]{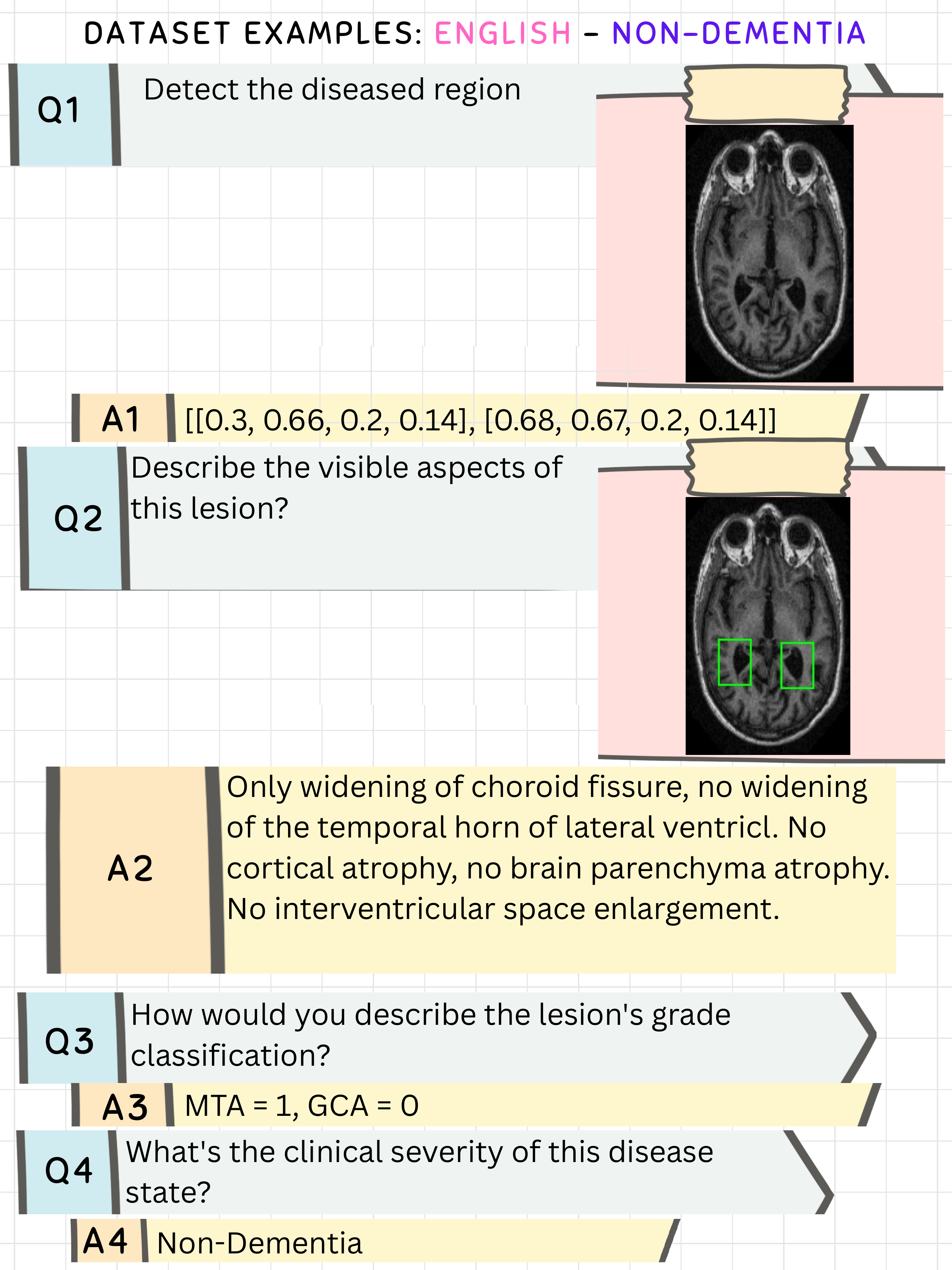}
        \caption{\textcolor{magenta}{\textbf{English}} – \textcolor{blue}{\textbf{Non-dementia}}}
        \label{fig:datasetexamples_english_nondementia}
    \end{subfigure}
    \hfill
    \begin{subfigure}{0.48\linewidth}
        \includegraphics[width=\linewidth]{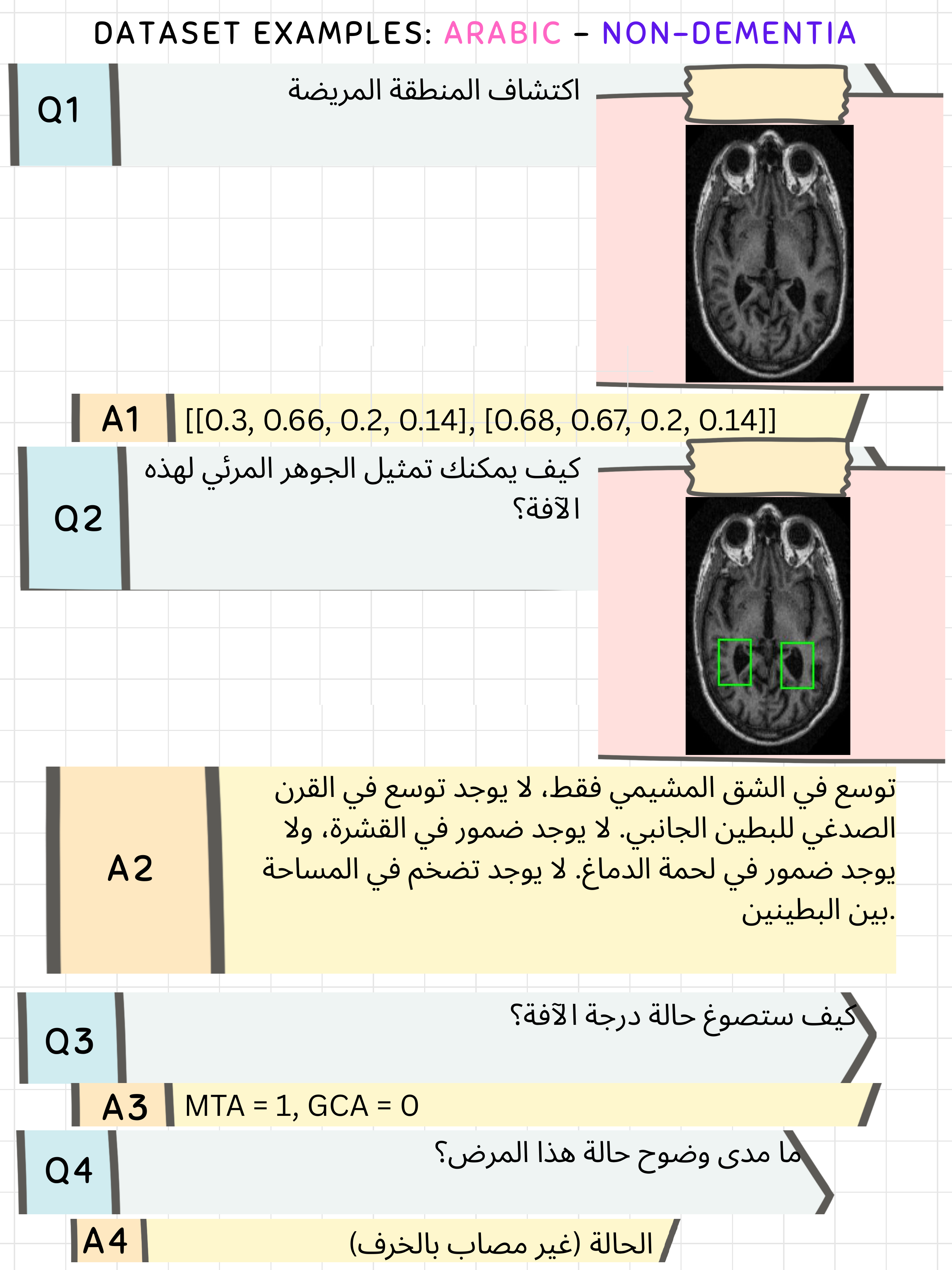}
        \caption{\textcolor{magenta}{\textbf{Arabic}} – \textcolor{blue}{\textbf{Non-dementia}}}
        \label{fig:datasetexamples_arabic_nondementia}
    \end{subfigure}

    \vspace{0.5cm}

    \begin{subfigure}{0.48\linewidth}
        \includegraphics[width=\linewidth]{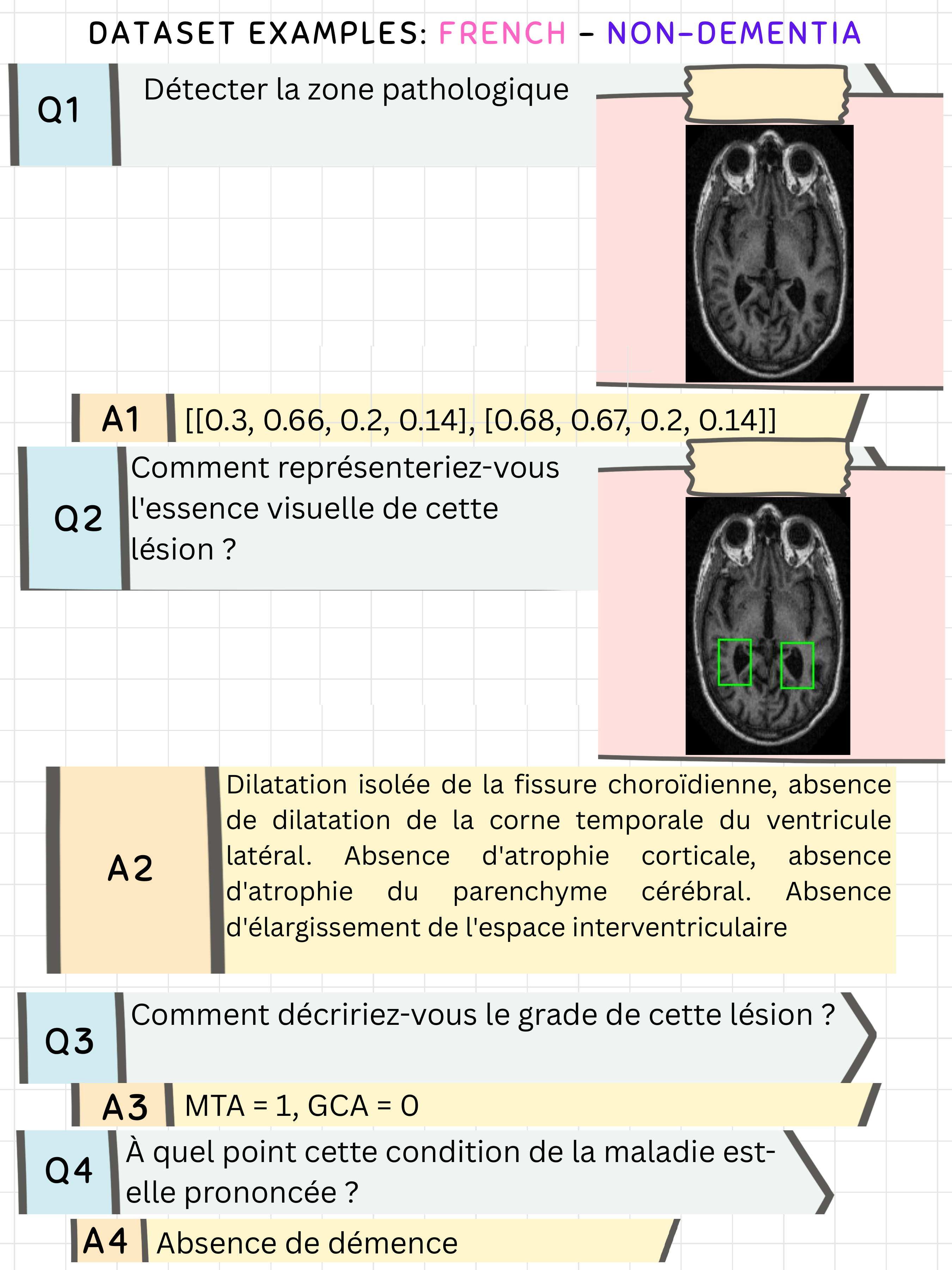}
        \caption{\textcolor{magenta}{\textbf{French}} – \textcolor{blue}{\textbf{Non-dementia}}}
        \label{fig:datasetexamples_french_nondementia}
    \end{subfigure}
    \hfill
    \begin{subfigure}{0.48\linewidth}
        \includegraphics[width=\linewidth]{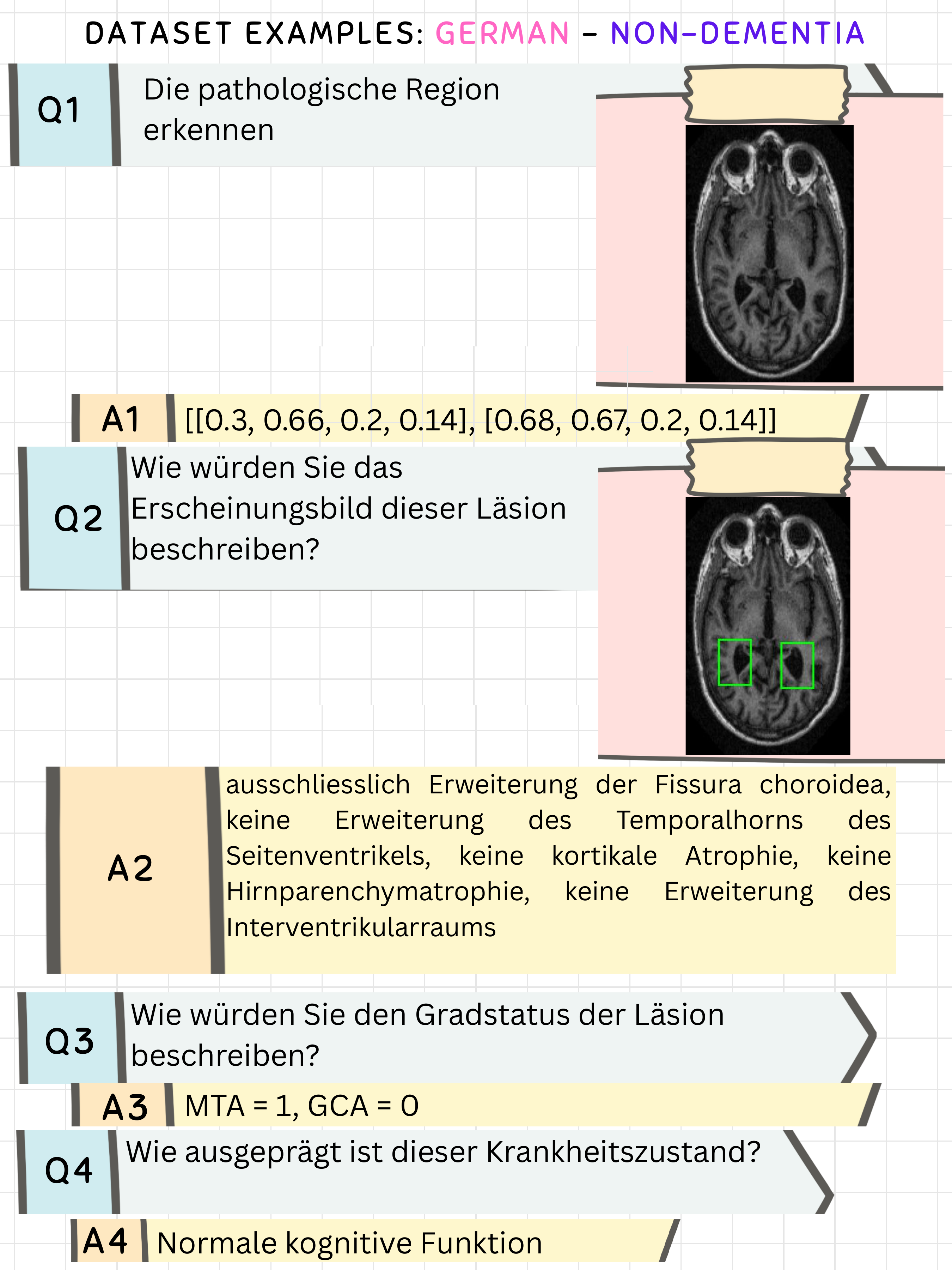}
        \caption{\textcolor{magenta}{\textbf{German}} – \textcolor{blue}{\textbf{Non-dementia}}}
        \label{fig:datasetexamples_german_nondementia}
    \end{subfigure}

    \caption{\textbf{Dataset examples} in the form of multi-turn \gls{VQA} conversations across four languages.  
    Each panel shows: \textcolor{magenta}{\textbf{Language}} (English, Arabic, French, German) and the diagnosis label \textcolor{blue}{\textbf{Non-dementia}}.  
    \\\textcolor{red}{\faUndo} \textcolor{red}{\textbf{Click back to:}} Section \ref{sec:dataset_examples} (\nameref{sec:dataset_examples}) or Table of \nameref{sec:table_of_content}.}
    \label{fig:datasetexamples_multilang_nondementia}
\end{figure}

Dataset examples for \textbf{Non-Dementia} follow this order: English (Figure \ref{fig:datasetexamples_english_nondementia}), Arabic (Figure \ref{fig:datasetexamples_arabic_nondementia}), French (Figure \ref{fig:datasetexamples_french_nondementia}), German (Figure \ref{fig:datasetexamples_german_nondementia}).

\begin{figure}[H]
    \centering
    \begin{subfigure}{0.48\linewidth}
        \includegraphics[width=\linewidth]{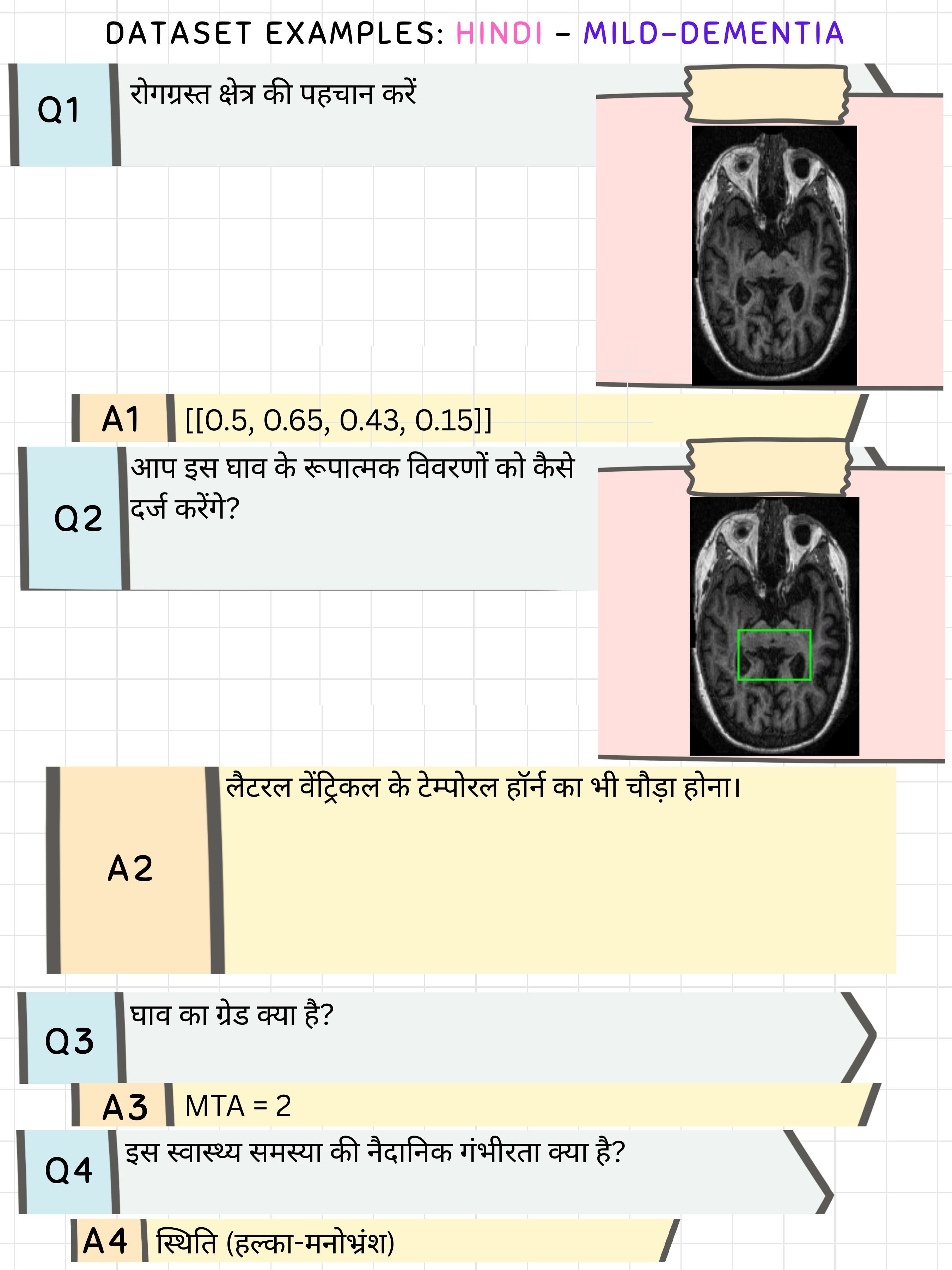}
        \caption{\textcolor{magenta}{\textbf{Hindi}} – \textcolor{blue}{\textbf{Mild-Dementia}}}
        \label{fig:datasetexamples_hindi_milddementia}
    \end{subfigure}
    \hfill
    \begin{subfigure}{0.48\linewidth}
        \includegraphics[width=\linewidth]{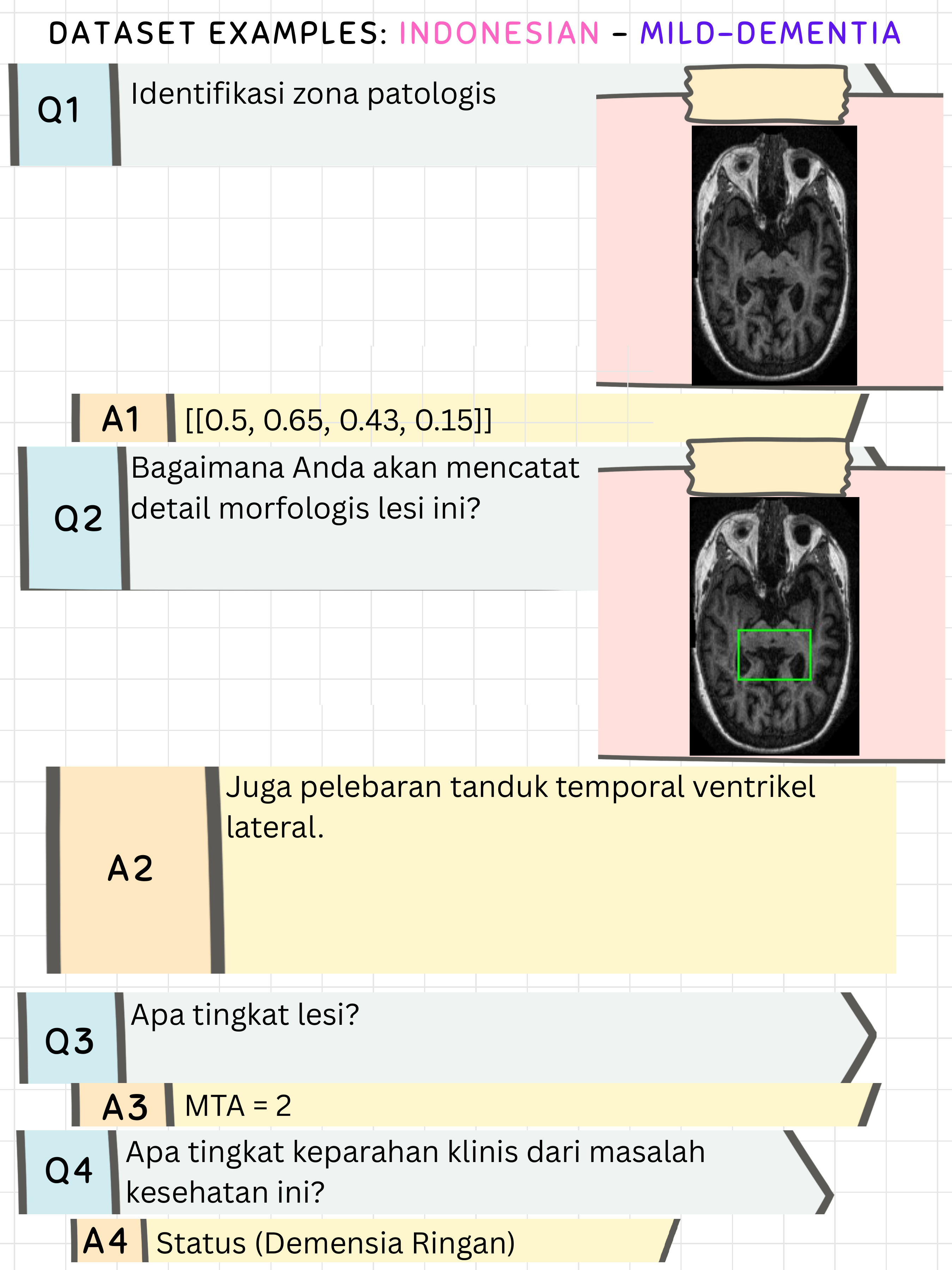}
        \caption{\textcolor{magenta}{\textbf{Indonesian}} – \textcolor{blue}{\textbf{Mild-Dementia}}}
        \label{fig:datasetexamples_indonesian_milddementia}
    \end{subfigure}

    \vspace{0.5cm}

    \begin{subfigure}{0.48\linewidth}
        \includegraphics[width=\linewidth]{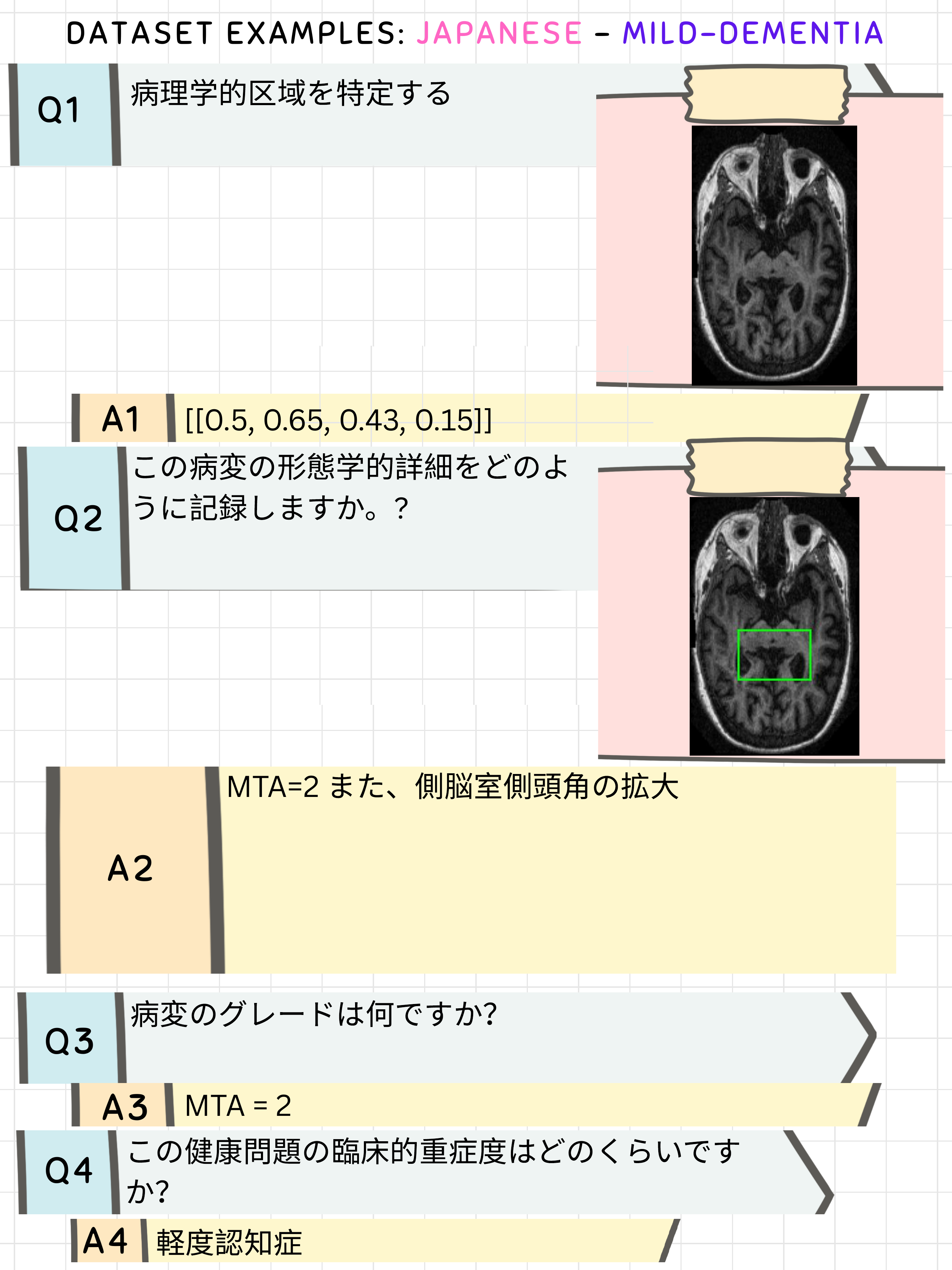}
        \caption{\textcolor{magenta}{\textbf{Japanese}} – \textcolor{blue}{\textbf{Mild-Dementia}}}
        \label{fig:datasetexamples_japanese_milddementia}
    \end{subfigure}
    \hfill
    \begin{subfigure}{0.48\linewidth}
        \includegraphics[width=\linewidth]{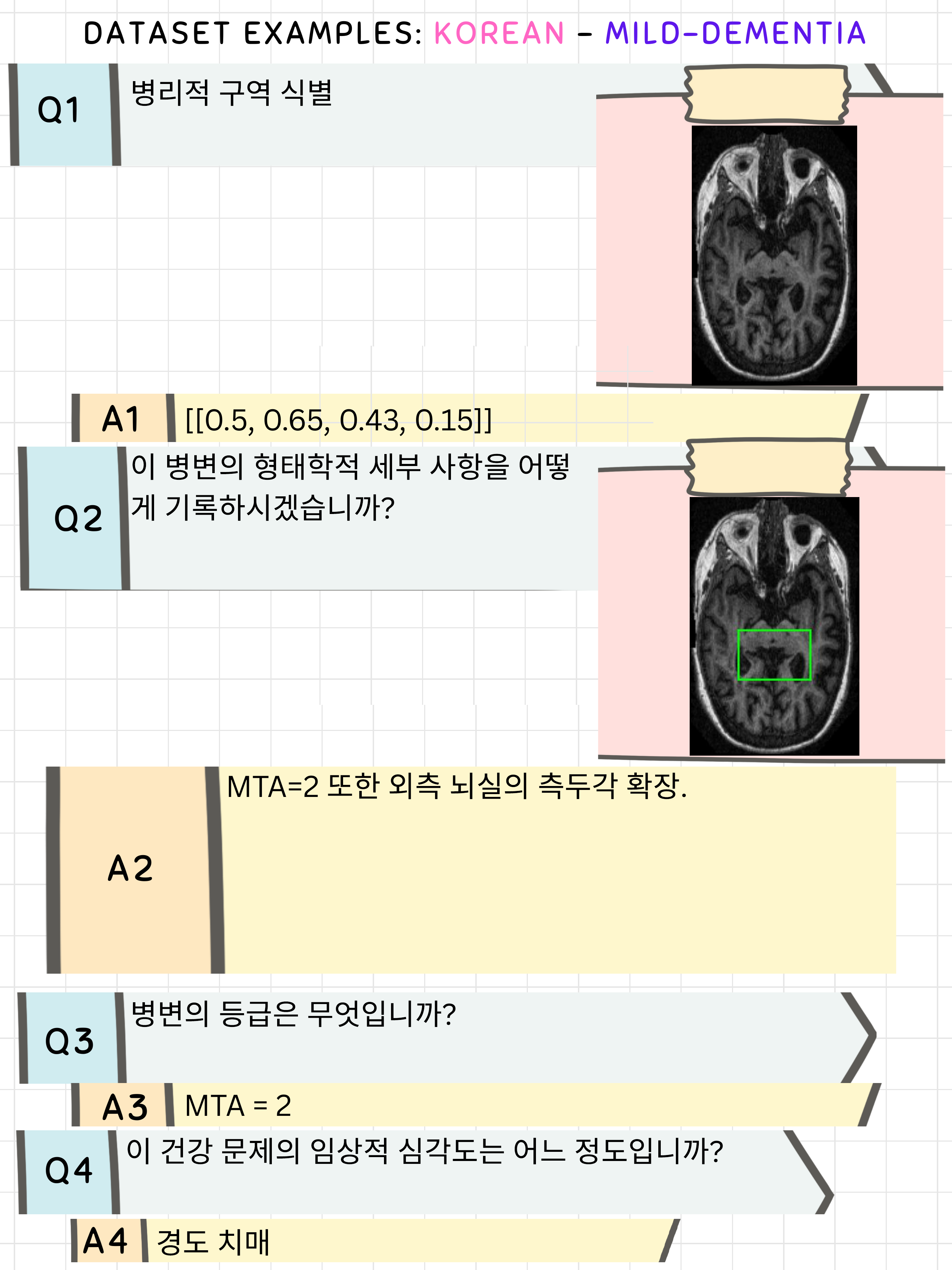}
        \caption{\textcolor{magenta}{\textbf{Korean}} – \textcolor{blue}{\textbf{Mild-Dementia}}}
        \label{fig:datasetexamples_korean_milddementia}
    \end{subfigure}

    \caption{\textbf{Dataset examples} in the form of multi-turn \gls{VQA} conversations across four languages.  
    Each panel explicitly shows: \textcolor{magenta}{\textbf{Language}} (Hindi, Indonesian, Japanese, Korean) and the diagnosis label \textcolor{blue}{\textbf{Mild-Dementia}}.  
    \\\textcolor{red}{\faUndo} \textcolor{red}{\textbf{Click back to:}} Section \ref{sec:dataset_examples} (\nameref{sec:dataset_examples}) or Table of \nameref{sec:table_of_content}.}
    \label{fig:datasetexamples_multilang_milddementia}
\end{figure}

Dataset examples for \textbf{Mild-Dementia} follow this order: Hindi (Figure \ref{fig:datasetexamples_hindi_milddementia}), Indonesian (Figure \ref{fig:datasetexamples_indonesian_milddementia}), Japanese (Figure \ref{fig:datasetexamples_japanese_milddementia}), Korean (Figure \ref{fig:datasetexamples_korean_milddementia}).

\begin{figure}[H]
    \centering
    \begin{subfigure}{0.48\linewidth}
        \includegraphics[width=\linewidth]{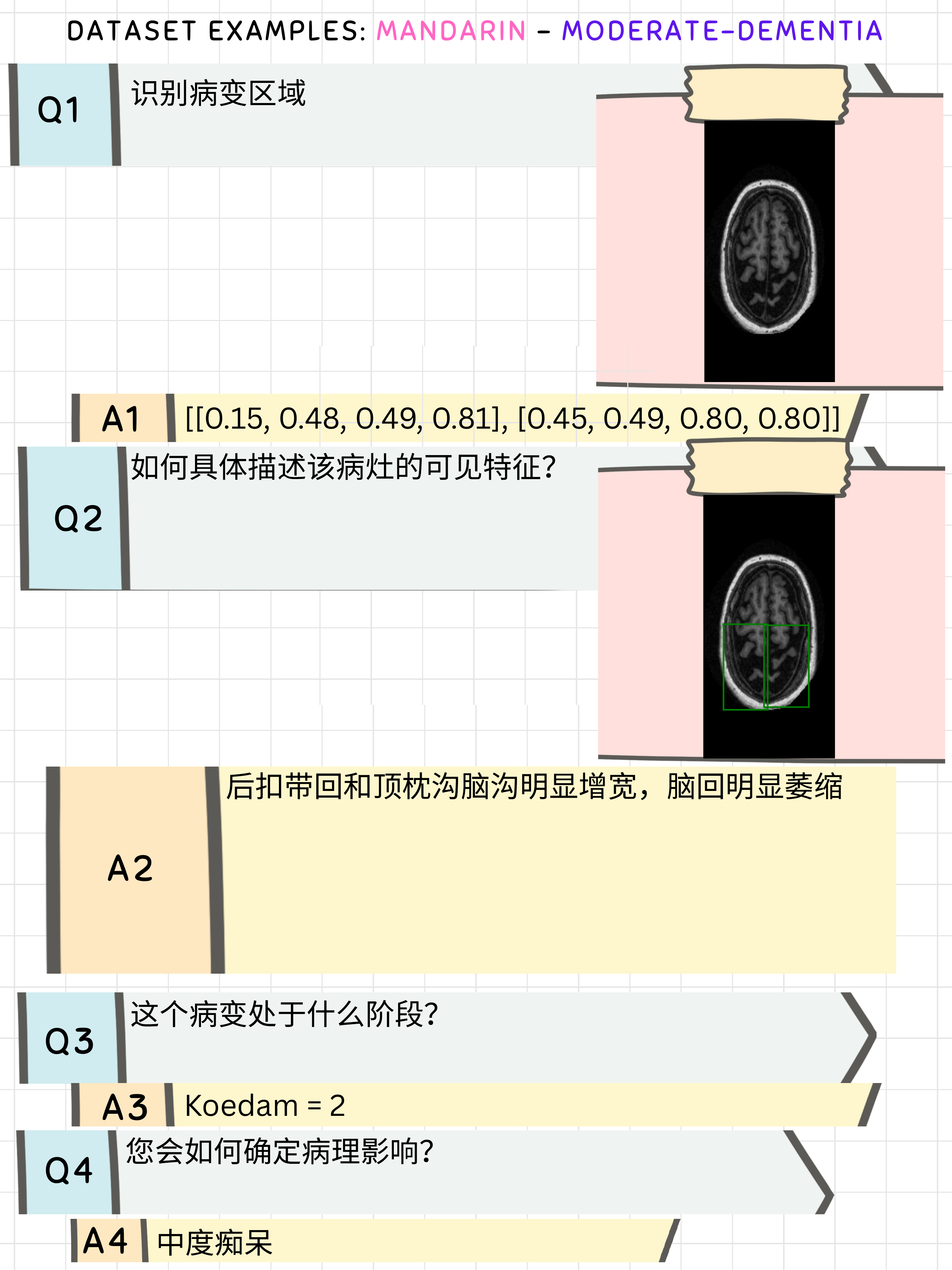}
        \caption{\textcolor{magenta}{\textbf{Mandarin}} – \textcolor{blue}{\textbf{Moderate-Dementia}}}
        \label{fig:datasetexamples_mandarin_moderatedementia}
    \end{subfigure}
    \hfill
    \begin{subfigure}{0.48\linewidth}
        \includegraphics[width=\linewidth]{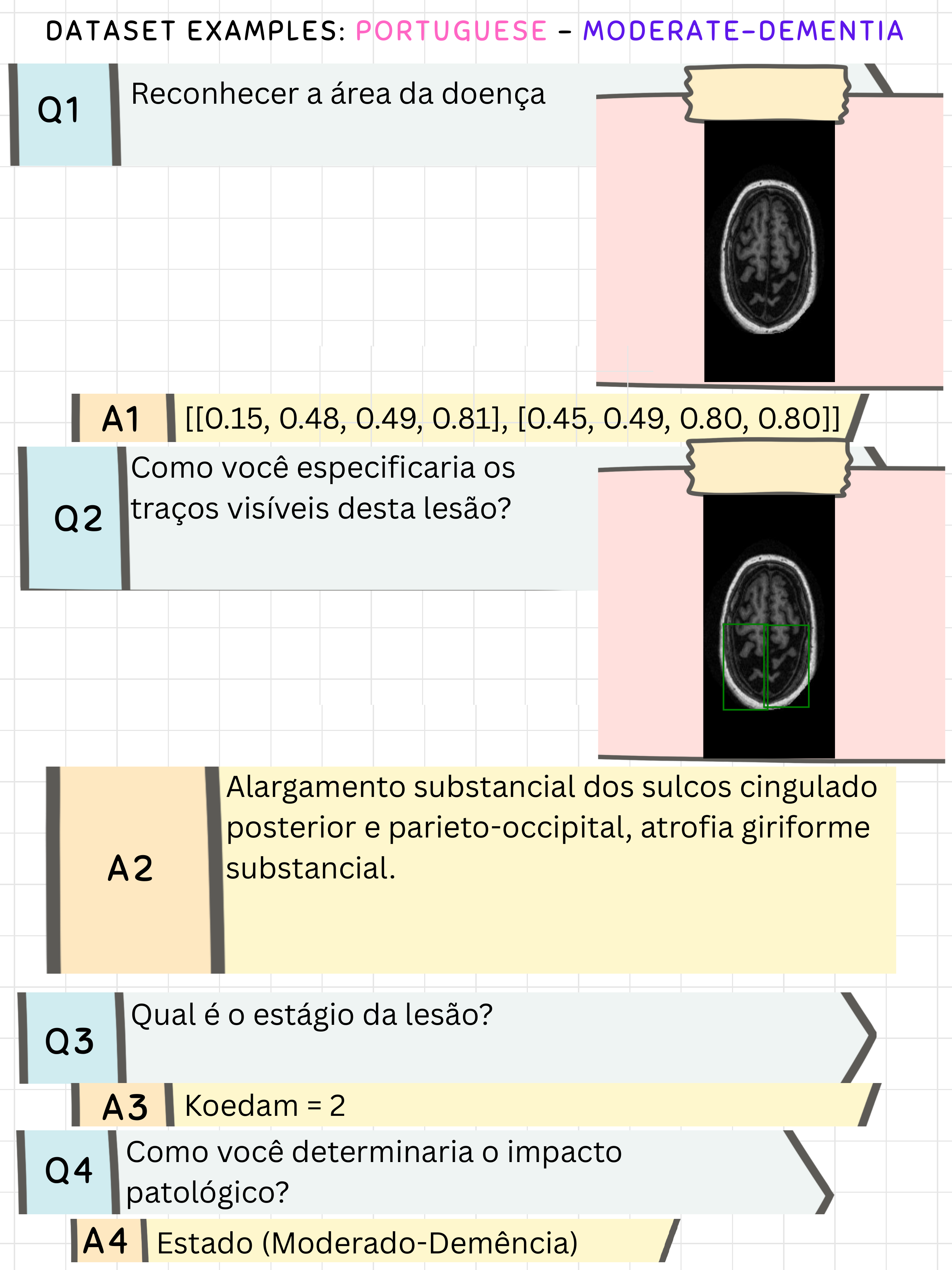}
        \caption{\textcolor{magenta}{\textbf{Portuguese}} – \textcolor{blue}{\textbf{Moderate-Dementia}}}
        \label{fig:datasetexamples_portuguese_moderatedementia}
    \end{subfigure}

    \vspace{0.5cm}

    \begin{subfigure}{0.48\linewidth}
        \includegraphics[width=\linewidth]{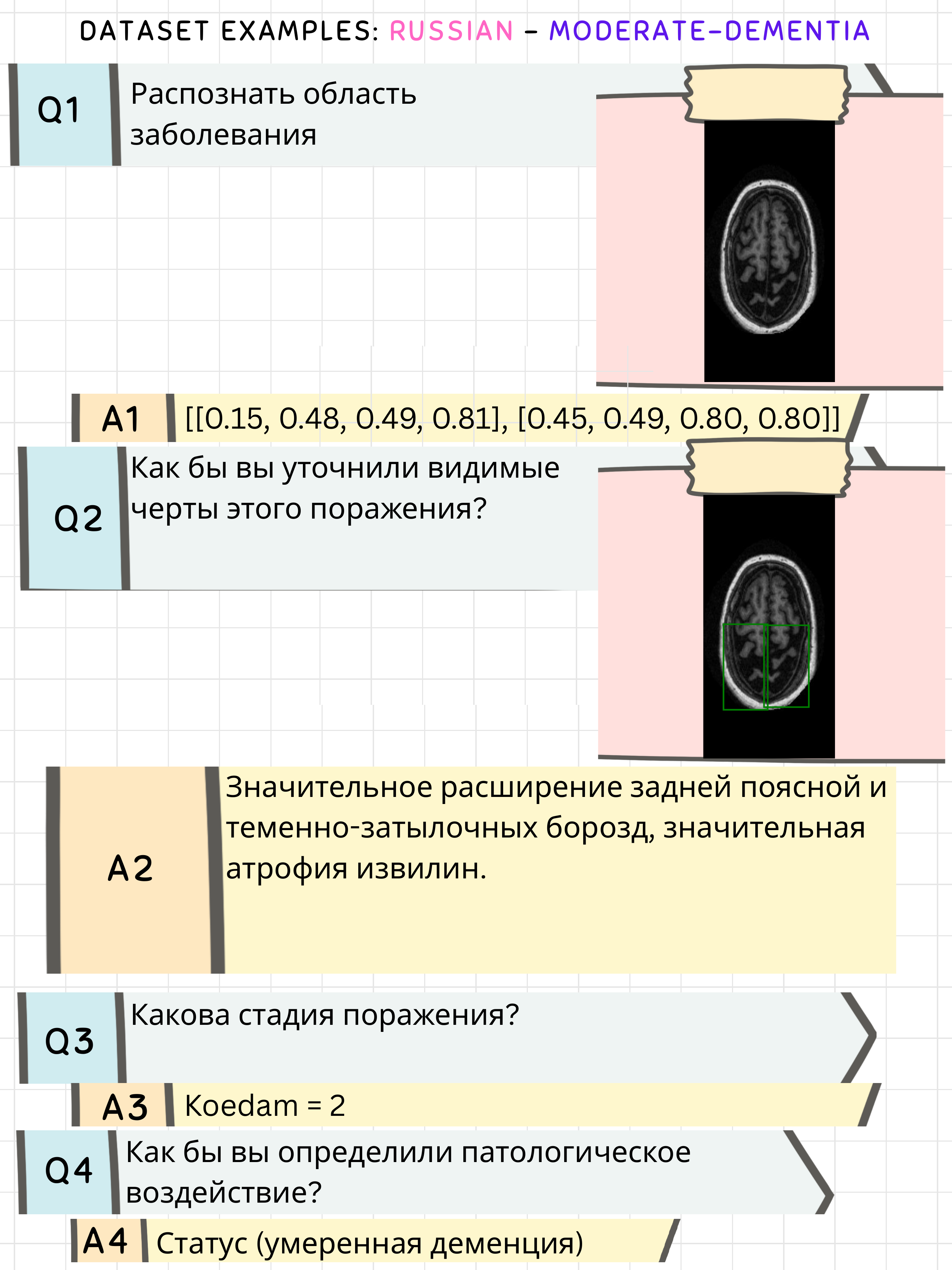}
        \caption{\textcolor{magenta}{\textbf{Russian}} – \textcolor{blue}{\textbf{Moderate-Dementia}}}
        \label{fig:datasetexamples_russian_moderatedementia}
    \end{subfigure}
    \hfill
    \begin{subfigure}{0.48\linewidth}
        \includegraphics[width=\linewidth]{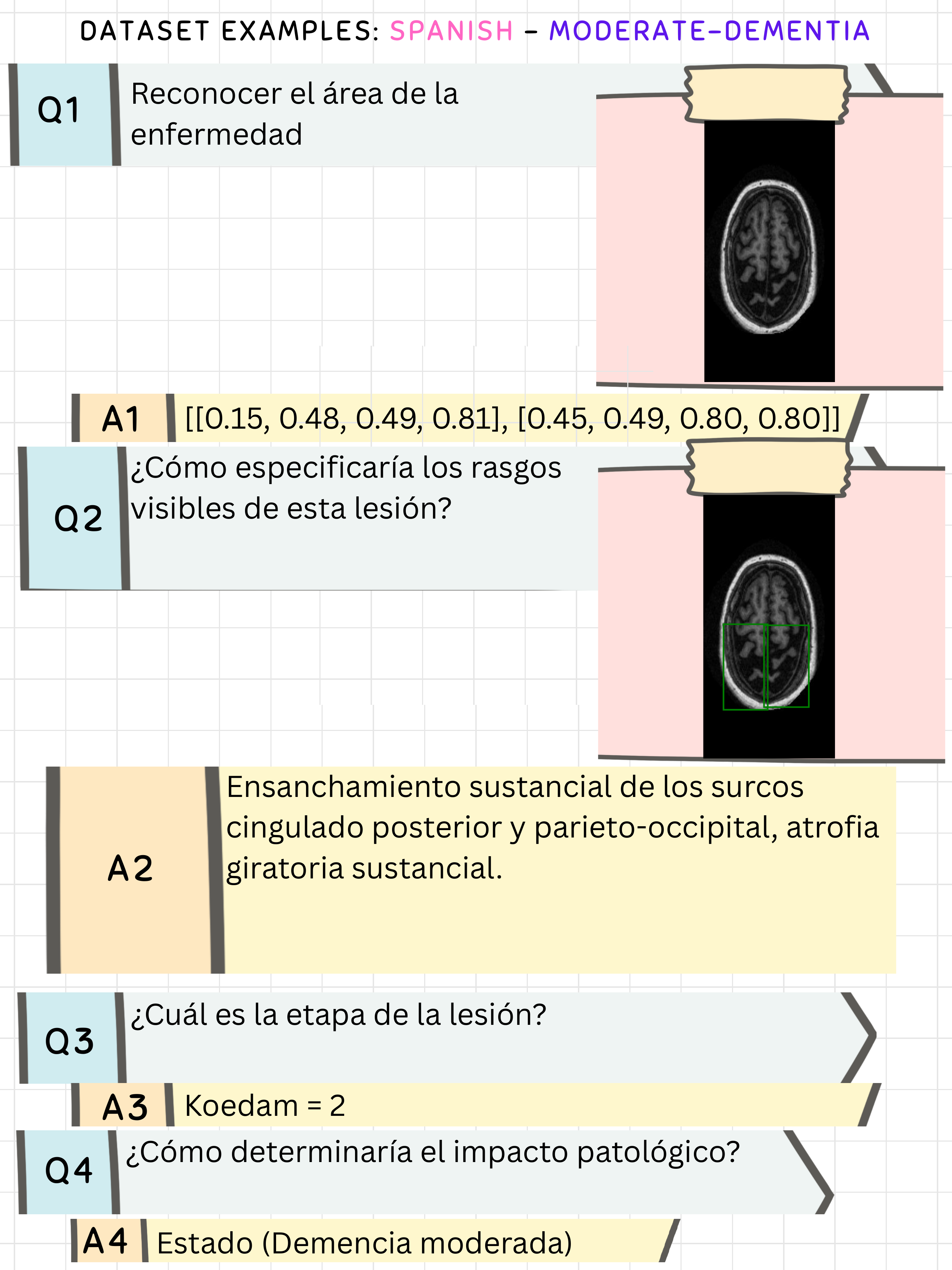}
        \caption{\textcolor{magenta}{\textbf{Spanish}} – \textcolor{blue}{\textbf{Moderate-Dementia}}}
        \label{fig:datasetexamples_spanish_moderatedementia}
    \end{subfigure}

    \caption{\textbf{Dataset examples} in the form of multi-turn \gls{VQA} conversations across four languages.  
    Each panel shows: \textcolor{magenta}{\textbf{Language}} (Mandarin, Portuguese, Russian, Spanish) and the diagnosis label \textcolor{blue}{\textbf{Moderate-Dementia}}.  
    \\\textcolor{red}{\faUndo} \textcolor{red}{\textbf{Click back to:}} Section \ref{sec:dataset_examples} (\nameref{sec:dataset_examples}) or Table of \nameref{sec:table_of_content}.}
    \label{fig:datasetexamples_multilang_moderatedementia}
\end{figure}

Dataset examples for \textbf{Moderate-Dementia} follow this order: Mandarin (Figure \ref{fig:datasetexamples_mandarin_moderatedementia}), Portuguese (Figure \ref{fig:datasetexamples_portuguese_moderatedementia}), Russian (Figure \ref{fig:datasetexamples_russian_moderatedementia}), Spanish (Figure \ref{fig:datasetexamples_spanish_moderatedementia}).

\subsection{S-Chain Dataset Comparison with Other General Visual CoT}
As shown in Table~\ref{tab:cot_data_compare}, \textbf{S-Chain is one of the largest visual \gls{CoT} datasets to date}, with 197k examples (172k train/val, 25k test combined multi-lingual). Unlike general visual \gls{CoT} datasets, it uniquely combines stepwise reasoning with explicit region-level grounding, supporting large-scale evaluation of both interpretability and diagnostic accuracy beyond final answers.

\begin{table}[H]
\centering
\vspace{-0.05in}
\scalebox{0.95}{
\renewcommand{\arraystretch}{1.1}
\setlength{\tabcolsep}{6pt}
\begin{tabular}{l|cc|ccc}
\toprule
\rowcolor{headerorange}
\textbf{Dataset} & \textbf{Train+Val} & \textbf{Test} & \textbf{CoT} & \textbf{Grounding} & \textbf{Expert Annotation} \\
\midrule
Visual7W~\citep{zhu2016visual7w} & 229,557 & 98,382 &  & \textcolor{green}{\ding{51}} & \\
ScienceQA~\citep{lu2022learn} & 16,967 & 4,241 & \textcolor{green}{\ding{51}} & & \\
MME-CoT~\citep{jiangmme} & -- & 1,130 & \textcolor{green}{\ding{51}} & & \\
MM-GCoT~\citep{wu2025grounded} & 23,028 & 994 & \textcolor{green}{\ding{51}} & \textcolor{green}{\ding{51}} & \\
\midrule
\SChain \, \textbf{(ours)} & 172,528 & 24,672 & \textbf{\textcolor{green}{\ding{51}}} & \textbf{\textcolor{green}{\ding{51}}} & \textbf{\textcolor{green}{\ding{51}}} \\
\bottomrule
\end{tabular}}
\caption{\textbf{Comparison between S-Chain and \textit{general} Visual \gls{CoT} datasets.}}
\label{tab:cot_data_compare}
\end{table}

\newpage

\newpage

\newpage

\onecolumn
\section{Annotation Guidelines}
\label{sec:annotation_guidelines}
\subsection{Clinical Motivation}
\subsubsection{Introduction to Alzheimer}
\gls{AD} is the most common type of dementia, accounting for an estimated 60\% to 80\% of dementia among individuals aged 65 and older. It is also listed as the world's fifth most common cause of death \cite{kumar2024alzheimer, trinh2024impact}. The lifetime risk of developing \gls{AD} at age 45 is 1 in 5 for women and 1 in 10 for men. \gls{AD} is a chronic, progressive neurodegenerative disorder clinically characterized by progressive memory loss with functional impairments in the frontal/executive, visuospatial, and language domains. 

Pathologically, this disease is characterized by the accumulation of \gls{ABeta} plaques and \gls{NFT} in the brain, as well as synapse loss and neurodegeneration \cite{long2023identifying, rajmohan2017amyloid}. Histopathological findings include accumulating \gls{ABeta} plaques, synaptic loss in \gls{NFT}, and neurodegeneration \cite{apostolova2016alzheimer, he2022design, hampel2021developing}. 

To date, \gls{AD} remains a disease with no specific cure. Therefore, the goal of further improvement in diagnosis is early diagnosis, which stems from this reason as well as the prevalence of the above-mentioned related pathologies. Especially in developed countries with a predominance of elderly populations. 

Today, the diagnosis and follow-up of all neurodegenerative diseases cannot be performed without radiological imaging, primarily \gls{MRI}, \gls{PET} \cite{jeong2021restoration, chappell2021partial}. Although \gls{PET} serves as the gold standard for diagnosing \gls{AD}, it is significantly higher than \gls{MRI}. However, the cost is also many times higher than \gls{MRI}. The economic burden is very large for patients because this is a chronic disease, requiring frequent follow-up and repeat examinations of imaging tests.

\textbf{For this reason, we decided to establish this study on \gls{MRI} for better diagnosis and monitoring in patients with dementia.} Simultaneously using several different semiquantitative scales has been designed to improve the precision of assessment and reduce inter-observer variability.

\subsubsection{Rationale}
Early and accurate diagnosis of \gls{AD} remains a major clinical challenge, especially during the prodromal and \gls{MCI} stages when therapeutic interventions may be most beneficial. Although biomarkers such as \gls{CSF} analysis and \gls{PET} imaging have improved diagnostic precision, their high cost, invasiveness, and limited availability restrict their routine clinical use, particularly in low-resource settings \cite{sanaat2023cycle}. Consequently, there is a growing need for accessible, non-invasive, and cost-effective diagnostic tools, with structural \gls{MRI} being one of the most practical and widely available options.

Recent studies have demonstrated that specific regional patterns of brain atrophy, observable on \gls{MRI}, strongly correlate with underlying \gls{AD} pathology. In particular, visual rating scales such as the \gls{MTA} scale, the \gls{GCA} scale, and the \gls{Koedam} for posterior atrophy have been increasingly adopted in both clinical and research settings. These tools offer a semiquantitative approach to assessing structural changes and are valuable for distinguishing \gls{AD} from other dementias such as \gls{FTD} or \gls{DLB} \cite{ferreira2015practical, chouliaras2023use}.

Several recent studies support the clinical relevance and diagnostic performance of these scales. For example, the Scheltens \gls{MTA} scale has been shown to correlate well with hippocampal volumetry and reliably distinguish \gls{AD} patients from healthy controls \cite{maartensson2020medial, molinder2021validity}. Likewise, the \gls{Koedam} has demonstrated utility in identifying early-onset or atypical \gls{AD} presentations with posterior atrophy patterns \cite{fumagalli2020parieto, graff2021new}. However, each scale individually has limitations in sensitivity, especially in early or mixed pathology cases. Therefore, combining multiple scales may enhance diagnostic accuracy and provide a more comprehensive structural assessment of the brain \cite{bruun2018evaluating}.

Our clinical study aims to build on this body of evidence by implementing a standardized annotation protocol using all three visual rating scales across a diverse patient cohort. By doing so, we hope to reduce inter-rater variability, improve early detection, and establish a robust \gls{MRI}-based framework that can support \gls{AI}-assisted diagnosis and longitudinal monitoring of \gls{AD}.

\subsection{Etiology}
\subsubsection{Molecular Pathology and Protein Aggregation}
Two hallmark protein abnormalities at the core of \gls{AD} pathology are extracellular deposition of \gls{ABeta} plaques and intracellular accumulation of hyperphosphorylated tau protein, forming \gls{NFT} \cite{zhang2021interaction}. The amyloid cascade hypothesis proposes that the overproduction or impaired clearance of \gls{ABeta} peptides, particularly \gls{ABeta}, initiates a cascade of events including synaptic dysfunction, tau pathology, neuroinflammation, and ultimately neuronal death. Tau pathology, while also found in other tauopathies, becomes pathogenic in \gls{AD} when it spreads in a stereotypical pattern across vulnerable brain regions, particularly the hippocampus and entorhinal cortex \cite{zhang2023amyloid}.

\subsubsection{Neuroinflammation and Microglial Dysfunction}
Microglia, the resident immune cells of the brain, play a dual role in \gls{AD}. Initially, they attempt to clear misfolded proteins through phagocytosis. However, in the presence of chronic \gls{ABeta} accumulation, microglia can shift toward a pro-inflammatory state, releasing cytokines that exacerbate neuronal damage \cite{miao2023microglia}. Genetic studies have highlighted the importance of microglial function in \gls{AD} pathogenesis, particularly through mutations in genes such as TREM2, which impair the microglial response and enhance vulnerability to disease \cite{qu2023microglial, li2023trem2}.

\subsubsection{Genetic Risk Factors}
Genetic susceptibility significantly contributes to \gls{AD} risk, particularly in early-onset familial cases, which are often linked to autosomal dominant mutations in genes such as APP, PSEN1, and PSEN2 \cite{bekris2010genetics}. In late-onset \gls{AD}, the most well-established genetic risk factor is the $\epsilon$4 allele of the \gls{APOE} gene \cite{montufar2017association}. Carriers of one or two copies of the \gls{APOE}-$\epsilon$4 allele have an increased risk and earlier onset of the disease, likely due to reduced clearance of \gls{ABeta} and heightened inflammatory responses. Other genetic loci, including CLU, PICALM, CR1, and rare TREM2 variants (e.g., R47H), also modulate risk through pathways related to lipid metabolism, synaptic function, and immune regulation \cite{karch2015alzheimer}.

\subsubsection{Environmental and Lifestyle Factors}
While genetics plays a foundational role, modifiable risk factors are increasingly recognized in \gls{AD} pathogenesis. These include cardiovascular risk factors such as hypertension, diabetes, obesity, and hyperlipidemia, which may compromise cerebral perfusion and exacerbate neurodegeneration. Lifestyle-related factors such as low educational attainment, social isolation, physical inactivity, smoking, and poor diet have also been linked to increased \gls{AD} risk, possibly by reducing cognitive reserve and promoting systemic inflammation \cite{santos2017pathophysiologic, edwards2019modifiable}.

\subsubsection{Age and Comorbidities}
Age remains the strongest non-modifiable risk factor for \gls{AD}, with prevalence doubling approximately every five years after the age of 65 \cite{kumar2024b_alzheimer}. The aging brain undergoes several changes that may predispose it to \gls{AD} pathology, including mitochondrial dysfunction, oxidative stress, impaired proteostasis, and reduced synaptic plasticity. Moreover, comorbid conditions such as cerebrovascular disease, depression, and traumatic brain injury can interact with underlying \gls{AD} pathology to influence clinical presentation and progression \cite{kumar2024alzheimer}.

Understanding the etiology of \gls{AD} is essential for interpreting structural and functional brain changes observed on \gls{MRI}. The progressive accumulation of \gls{ABeta} and hyperphosphorylated tau proteins, key pathological hallmarks of \gls{AD}, leads to synaptic loss, neuronal degeneration, and brain atrophy-changes that are detectable with \gls{MRI}. Structural \gls{MRI} is particularly sensitive to the neurodegenerative effects of these pathological processes, revealing region-specific atrophy patterns. The medial temporal lobe, including the hippocampus, entorhinal cortex, and parahippocampal gyrus, is typically affected in the early stages of \gls{AD} due to its vulnerability to tau pathology \cite{ravikumar2024postmortem, vemuri2010role}. 

As the disease progresses, atrophy extends to the parietal and frontal lobes. These imaging patterns reflect the underlying etiology and provide supportive evidence for diagnosis and staging. Moreover, advanced \gls{MRI} techniques such as \gls{DTI} and volumetric analysis offer insights into white matter integrity and brain network disintegration, which are indirectly linked to protein aggregation, neuroinflammation, and genetic risk factors (e.g., \gls{APOE} $\epsilon$4 status). Thus, \gls{MRI} serves as a bridge between the biological mechanisms of \gls{AD} and clinical decision-making, enabling early detection, differential diagnosis, and monitoring of disease progression \cite{monica2021alzheimer}.
\subsection{Pathophysiology}
\gls{AD}, like other neurodegenerative dementias, follows a gradually progressive course marked by the accumulation of misfolded proteins in the brain. These abnormal proteins—primarily \gls{ABeta} and tau-disrupt normal cellular processes and initiate a cascade of pathological changes. In many cases with each proteinopathy contributes to distinct clinical phenotypes \cite{monica2021alzheimer}. The formation of these toxic aggregates is believed to result from an imbalance between protein production and clearance mechanisms. In response, the brain's innate immune cells, microglia, become activated and initiate protective responses aimed at repair and removal. However, persistent protein accumulation can drive microglia into a pro-inflammatory state, shifting from an acute, self-limiting process to chronic neuroinflammation-an event central to ongoing neuronal injury \cite{allegri2020moving}.

Advancements in molecular imaging and pathology have highlighted overlaps between neurodegenerative phenotypes. Although tau protein aggregation is observed in \gls{AD}, the disease is not considered a primary tauopathy due to the dominant role of \gls{ABeta} pathology. The characteristic spatial progression of tau and \gls{ABeta} accumulation aligns with the atrophy patterns seen in structural \gls{MRI}, particularly affecting the hippocampus and adjacent medial temporal lobe structures in early stages \cite{sengupta2022amyloid}.

The pathophysiological process is influenced by a combination of genetic predispositions and environmental exposures. Non-modifiable risk factors include advancing age and inherited genetic variants. Among the most recognized genetic contributors is the \gls{APOE} $\epsilon$4 allele, where homozygous carriers are at significantly elevated risk for developing \gls{AD}. Another important genetic factor involves rare mutations in the triggering receptor expressed on myeloid cells 2 (TREM2) gene \cite{montufar2017association}. Depending on the specific variant, such as R47H or R62H, microglial responses can range from neuroprotective to dysfunctional, impairing the clearance of pathological proteins and worsening disease progression \cite{karch2015alzheimer}.

Conversely, several modifiable risk factors have been identified and offer potential avenues for prevention and risk reduction. These include physical inactivity, tobacco use, limited education, reduced cognitive and social engagement, hypertension, diabetes mellitus, and poor dietary habits. These lifestyle-related factors are believed to influence brain resilience and may interact with underlying pathological processes to modify the trajectory of disease onset and progression.

Incorporating \gls{MRI} into the study of \gls{AD} pathophysiology provides a non-invasive window into these molecular and cellular changes, allowing for early detection of structural brain alterations that reflect the underlying disease mechanisms.

\subsection{Alzheimer Diagnosis}
The diagnosis of \gls{AD} remains primarily clinical, supported by cognitive testing, laboratory evaluations, and neuroimaging. According to the 2011 NIA-AA criteria and DSM-5, dementia is identified when cognitive or behavioral symptoms interfere with daily functioning, represent a decline from previous abilities, and are not better explained by psychiatric illness \cite{jack2018nia}. These deficits typically affect at least one domain, such as memory, executive function, language, visuospatial skills, or behavior.

Initial assessment includes a detailed medical history and mental status evaluation. A comprehensive assessment performed through a holistic evaluation that incorporated various factors, including clinical history, neuropsychological examination, cognitive evaluations using the Mini-Mental State Examination (MMSE), Clinical Dementia Rating Scale Sum of Boxes (CDR-SB), and Montreal Cognitive Assessment (MoCA) are widely used, administered by qualified physicians, laboratory findings, and \gls{MRI} \cite{mckhann2011diagnosis, arevalo2015mini, o2008staging, cedarbaum2013rationale}. Functional status is assessed through structured or informal evaluations of daily living activities. For \gls{AD} patients, MMSE scores ranged between 18 and 26 \cite{crum1993population, tiepolt2013influence}, and the CDR-SB scores were between 4.5 and 18 \cite{o2008staging, lynch2005clinical}. Patients were excluded from the study due to the following criteria: the presence of brain tumors, significant infarctions, and hemorrhages on the brain \gls{MRI} scan, and the patient's movement during \gls{PET} scanning.

Laboratory tests help exclude reversible causes, including metabolic, endocrine, or nutritional deficiencies. Additional investigations—such as \gls{CSF} analysis for \gls{ABeta} and tau biomarkers, Electroencephalogram (EEG), and genetic testing—may be indicated based on clinical context and availability.

Neuroimaging plays a key supportive role. \gls{MRI} is preferred over Computed Tomography (CT) for its superior sensitivity to early structural changes, particularly in the medial temporal lobe. It also helps exclude other causes, such as vascular lesions or tumors. Fluorodeoxyglucose (FDG)-\gls{PET} may reveal characteristic hypometabolism patterns, and while amyloid and tau \gls{PET} imaging offer more specific biomarker data, their clinical use is limited by accessibility and cost.

\subsection{MRI Findings}
\gls{MRI} is an essential tool in detecting structural brain changes associated with \gls{AD}. The distribution of affected areas in different entities explains the variation in symptoms and imaging patterns. Patterns of regional brain atrophy correlate with specific clinical symptoms and help differentiate \gls{AD} from other dementias. Three main visual rating scales-\gls{MTA}, \gls{GCA}, and \gls{Koedam} scores-are commonly used to assess characteristic atrophy patterns in \gls{AD}.

\subsubsection{Medial Temporal Lobe Atrophy (MTA)}
\begin{figure}[h]
    \centering
    \includegraphics[width=0.8\linewidth]{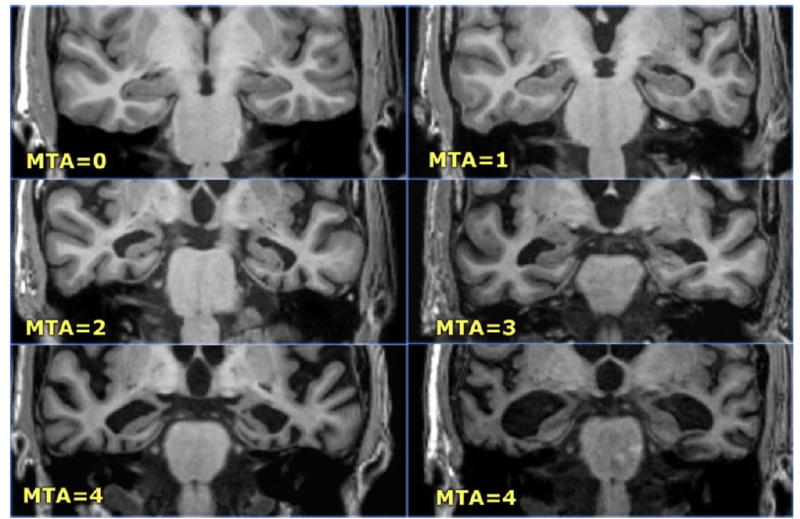}
    \caption{\textbf{Coronal T1W images show different degrees of medial temporal lobe atrophy in 5 different patients with Alzheimer’s disease clinical presentation.} Using the Scheltens scale, the medial temporal lobe is assessed on coronal planes: 
    \\(A) MTA 0 – normal width of the choroid fissure, the temporal horn, and a normal HC height; 
    \\(B) MTA 1 – mild widened choroid fissure, normal temporal horn, and HC height; 
    \\(C) MTA 2 – moderately widened choroid fissure, mild temporal horn enlargement, and a mild reduction in HC height; 
    \\(D) MTA 3 – markedly widened choroid fissure, a moderate enlargement of the temporal horn, and a moderate reduction in HC height; 
    \\(E) MTA 4 – markedly widened choroid fissure, enlargement of the temporal horn, and a reduction in HC height. 
    \\\gls{MTA}: Medial temporal lobe atrophy; HC: Hippocampus \cite{vzivanovic2023role}.}
    \label{fig:Schelten_scale_MTA}
\end{figure}

The medial temporal lobe is an early affected site for \gls{AD}-related neurodegeneration \cite{braak1991neuropathological}. \gls{MRI} can detect the regional atrophy of the medial temporal lobe structures, which is an essential \gls{AD} biomarker \cite{bobinski1999histological}. \gls{MTA} is one of the earliest and most prominent imaging features of \gls{AD}, typically involving the hippocampus, entorhinal cortex, and parahippocampal gyrus—regions essential for memory processing \cite{brinkmann2019segmentation}. The Scheltens \gls{MTA} scale is widely used in clinical practice to visually rate the degree of atrophy on coronal \gls{MRI} slices aligned perpendicular to the hippocampal axis \cite{scheltens1992atrophy, harper2015using}. It assesses three key features: hippocampal volume loss, widening of the choroid fissure, and enlargement of the temporal horn of the lateral ventricle, assigning scores from 0 (no atrophy) to 4 (severe atrophy), as shown in Table \ref{tab:Schelten_scale_MTA} and Figure \ref{fig:Schelten_scale_MTA}. While symmetrical atrophy is commonly seen in \gls{AD}, some asymmetry can occur. In our analysis, the dichotomized score of left and right was used. Early detection of hippocampal atrophy supports prodromal \gls{AD} diagnosis and helps differentiate it from other dementias such as \gls{FTD} and \gls{DLB}.

\begin{table}[h]
\centering
\renewcommand{\arraystretch}{1.5}
\resizebox{\textwidth}{!}{%
\begin{tabularx}{\textwidth}{>{\centering\arraybackslash}p{0.15\textwidth}|>{\raggedright\arraybackslash}X}
\hline
\textbf{MTA Score} & \textbf{Characteristics} \\
\hline
0 & Normal choroidal fissure width, temporal horn width, and HC volume. \\
1 & The choroidal fissure is mildly widened. \\
2 & Moderately widened choroidal fissure, minor temporal horn expansion of the lateral ventricle, and modest HC volume loss. \\
3 & Considerably expanded choroidal fissure, moderate temporal horn expansion, and moderate HC volume loss. \\
4 & Significantly expanded choroidal fissure, significantly enlarged temporal horn, and significantly reduced HC volume. \\
\hline
\multicolumn{2}{l}{\textit{$<$75 years: score 2 or more is abnormal.}} \\
\multicolumn{2}{l}{\textit{$>$75 years: score 3 or more is abnormal.}} \\
\multicolumn{2}{l}{\textbf{Abbreviations:} HC = Hippocampus; MTA = Medial temporal lobe atrophy.} \\
\hline
\end{tabularx}
}
\caption{\textbf{Scheltens scale for medial temporal lobe assessment (also known as \gls{MTA})}}
\label{tab:Schelten_scale_MTA}
\end{table}

Although hippocampal volumetry offers objective measurements, its accuracy can vary depending on the method used - manual tracing and different automated segmentation tools often delineate structures differently. In contrast, visual assessment using the \gls{MTA} scale remains more practical and reliable in routine clinical settings. Multiple studies have confirmed the \gls{MTA} scale’s ability to distinguish \gls{AD} patients from healthy controls, and comparisons with manual and automated volumetric methods have shown good to acceptable correlations \cite{susianti2024impact}.

A general problem with the \gls{MTA} score is the inconsistently defined cutoff value. Various cutoffs for pathological \gls{MTA} scores can be found in the literature, differing by age groups and education level. For example, Velickaite and colleagues elaborated that “at age 75, gender and education are confounders for \gls{MTA} grading. A score of $\ge 2$ is abnormal for low-educated women, and a score of $\ge 2.5$ is abnormal for men and highly educated women.” For this, the mean for both sides was considered together ((MTA score right $+$ MTA score left)$/$2) \cite{rau2021mta}.

\subsubsection{Global Cortical Atrophy (GCA)}
\begin{figure}[h]
    \centering
    \includegraphics[width=0.9\linewidth]{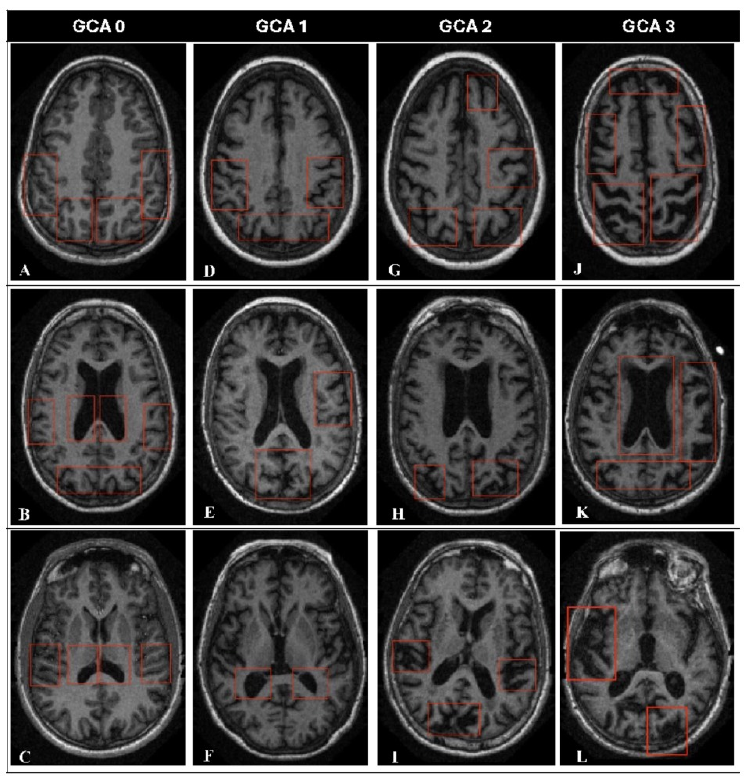}
    \caption{\textbf{Clinical Alzheimer's disease manifestation showing global cortical atrophy and ventricular dilatation in different stages.} 
    \\The first column shows: 
    \\(A) normal volume of the gyri and width of the sulci, 
    \\(B) normal dilatation of lateral ventricles, and 
    \\(C) normal dilatation of the third ventricle. 
    \\The second column shows: 
    \\(D) mild atrophy with a still normal volume of the gyri but some open sulci, 
    \\(E) mild dilatation of the lateral ventricles, 
    \\(F) mild dilatation of the third ventricle. 
    \\The third column shows: 
    \\(G) moderate brain atrophy with a reduction of gyri volume, and enlargement of the sulci, 
    \\(H) moderate dilatation of lateral ventricles, 
    \\(I) moderate dilatation of the third ventricle. 
    \\The fourth column shows: 
    \\(J) severe atrophy with severely reduced gyri, and enlarged sulci, 
    \\(K) severe dilatation of lateral ventricles, 
    \\(L) severe dilatation of the third ventricle. 
    \\The \textcolor{red}{red bounding boxes} are the signal of the \gls{GCA} scale.}
    \label{fig:GCA_scale}
\end{figure}

The \gls{GCA} scale, originally proposed by Pasquier, evaluates generalized cortical thinning across multiple brain regions, including the frontal, temporal, and parietal lobes. Each region is rated from 0 (normal) to 3 (severe atrophy) based on sulcal widening and gyral thinning, usually on axial FLAIR images, and detailed in Table \ref{tab:GCA_scale} and Figure \ref{fig:GCA_scale}. 

Total \gls{GCA} scores reflect the overall burden of brain atrophy. \gls{GCA} can be reliably classified on a semi-quantitative basis using standardized protocols and further quantified using volumetric analysis techniques \cite{al2018global}. Although \gls{GCA} can be influenced by normal aging, it becomes more significant in dementia when age-specific cutoffs are applied. Ventricular enlargement is also sometimes included to assess secondary atrophy, but it could be less specific for differentiating types of dementia.

\begin{table}[h]
\centering
\renewcommand{\arraystretch}{1.5}
\resizebox{\textwidth}{!}{%
\begin{tabular}{>{}c|>{\raggedright\arraybackslash}p{0.85\textwidth}}
\hline
\textbf{GCA} & \textbf{Characteristics} \\ \hline
0 & Normal volume of the gyri, sulci width, and ventricle dilatation; no cortical atrophy. \\
1 & Mild atrophy with still normal gyri volume, however with some slightly open sulci and mild ventricular dilatation. \\
2 & Moderate brain atrophy with reduced gyri volume, increased sulci, and moderate ventricular dilatation. \\
3 & Severe atrophy with significantly shrunken gyri, enlarged sulci, and dilated ventricles: ``knife blade''. \\ \hline
\multicolumn{2}{l}{\textit{GCA: Global cortical atrophy}} \\ \hline
\end{tabular}
}
\caption{\textbf{\gls{GCA}-scale for Global Cortical Atrophy}}
\label{tab:GCA_scale}
\end{table}

\subsubsection{Posterior Atrophy (Koedam Score)}
\begin{figure}[h]
    \centering
    \includegraphics[width=0.7\linewidth]{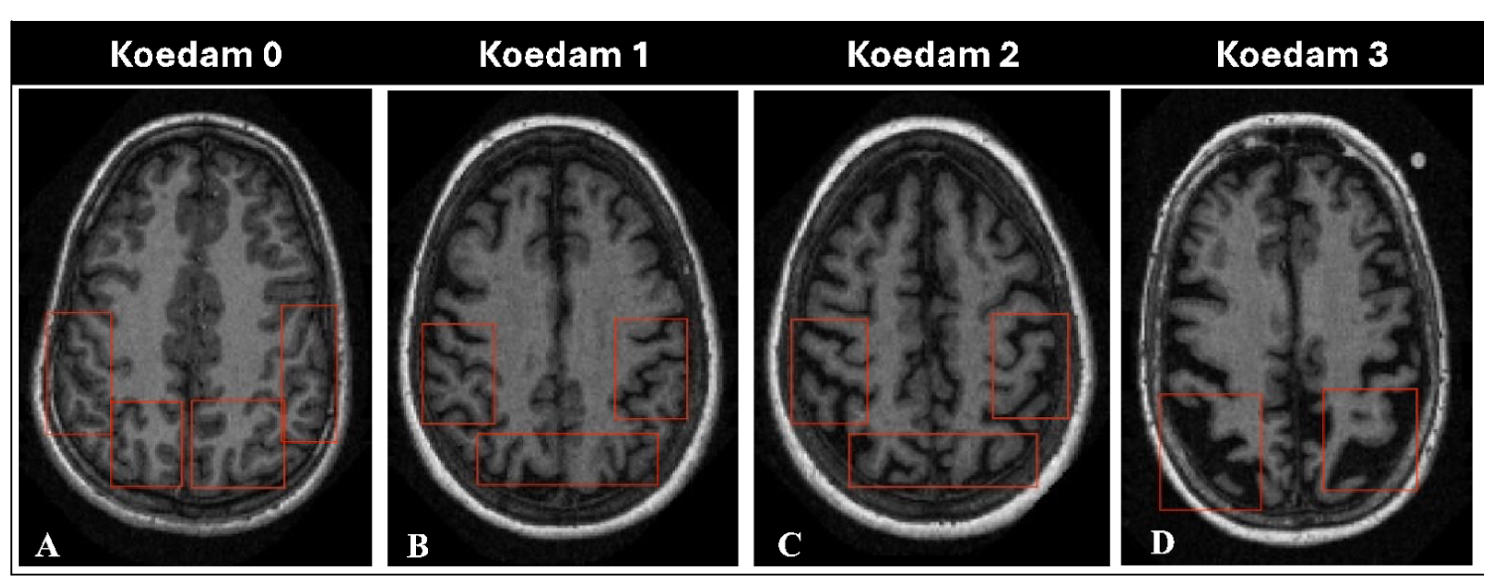}
    \caption{\textbf{Axial FLAIR, coronal T1W, and sagittal T1W images with Alzheimer’s disease show parietal atrophy scale.} 
    \\Koedam 0: (A) shows the closed posterior cingulate, parieto-occipital, and parietal lobe sulci.  
    \\Koedam 1: (B) mild posterior cingulate, parieto-occipital, and parietal lobe sulcal widening, and the mild atrophy of precuneus. 
    \\Koedam 2: (C) substantial posterior cingulate, parieto-occipital, and the parietal lobe sulcal widening, and substantial atrophy of precuneus. 
    \\Koedam 3: (D) extremal posterior cingulate, parieto-occipital, and the parietal lobe sulcal widening, and the mild atrophy of precuneus, knife-blade precuneus atrophy. 
    \\The \textcolor{red}{red bounding boxes}  are the signal of the \gls{Koedam} scale FLAIR: Three-dimensional T2-weighted fluid-attenuated inversion-recovery imaging.}
    \label{fig:koedam_scale}
\end{figure}

Posterior cortical atrophy is another important \gls{AD} imaging feature, especially in atypical forms. The \gls{Koedam} score provides a qualitative assessment of parietal atrophy based on sagittal \gls{MRI} views, as first described by \citet{koedam2011visual}, especially lobes, regions critical for visuospatial function. It assesses sulcal widening and cortical thinning across sagittal, axial, and coronal planes \cite{kaushik2020evaluation}, as shown in Figure \ref{fig:koedam_scale}. The score ranges from 0 (no atrophy) to 3 (severe atrophy) in each plane, as shown in Table \ref{tab:koedam_scale}. Posterior atrophy typically appears later in the disease course and can help differentiate \gls{AD} from other dementias, where posterior involvement is less prominent. According to \citet{yuan2019multiple}, the diagnostic performance of the \gls{Koedam} score is better in moderate and severe stages of \gls{AD} compared to mild cases \cite{yuan2019multiple}. Thus, incorporating the \gls{Koedam} score enhances the diagnostic accuracy, particularly in patients presenting with atypical or early-onset \gls{AD}.

\begin{table}[h]
\centering
\renewcommand{\arraystretch}{1.5}
\resizebox{\textwidth}{!}{%
\begin{tabular}{>{\centering\arraybackslash}m{1cm}| m{12cm}}
\toprule
\textbf{Score} & \textbf{Characteristics} \\
\midrule
0 & The posterior cingulate is closed, as also are the parieto-occipital sulcus, the parietal lobe sulci, and the precuneus. \\
1 & Mild posterior cingulate and parieto-occipital sulcus widening, with mild parietal lobe and precuneus atrophy. \\
2 & Significant expansion of the posterior cingulate and parieto-occipital sulcus, as well as significant atrophy of the parietal lobes and precuneus. \\
3 & End-stage atrophy with evident sulci expanding and knife-blade atrophy of the parietal lobes and precuneus. \\
\bottomrule
\end{tabular}
}
\caption{\textbf{\gls{Koedam} score for posterior atrophy assessment}}
\label{tab:koedam_scale}
\end{table}

In clinical practice, combining \gls{AD}, \gls{GCA}, and \gls{Koedam} offers a structured and efficient way to evaluate brain \gls{MRI} in patients with cognitive impairment. When interpreted alongside clinical history and neuropsychological testing, these imaging findings substantially improve diagnostic confidence and can support early and differential diagnosis of \gls{AD}.

\subsection{Method}
\subsubsection{Imaging acquisition}
All \gls{MRI} scans were acquired using a standardized protocol to ensure consistency and diagnostic quality. High-resolution T1-weighted images were obtained using a 3D magnetization-prepared rapid gradient echo (MPRAGE) sequence with the following typical parameters: repetition time (TR) $\approx$ 2,000 ms, echo time (TE) $\approx$ 2.5 ms, inversion time (TI) $\approx$ 900 ms, flip angle $\approx$ $9^\circ$, and voxel size $\approx$ 1 × 1 × 1 mm$^3$
. The acquisition was performed in the sagittal plane and included whole-brain coverage. Axial FLAIR and coronal T2-weighted images were also included to support the visual rating of cortical atrophy and to exclude other intracranial pathologies such as infarcts, tumors, or hydrocephalus. Images were visually inspected for quality, and scans with significant motion artifacts or structural abnormalities unrelated to neurodegeneration were excluded from the analysis.

\subsubsection{Annotation Protocol}
The annotation process was conducted in a stepwise manner by three specially trained physicians from three different institutions. Each expert independently reviewed the imaging data, beginning with the selection of the most representative slices from each patient. For each target brain region, four to five slices showing the clearest anatomical features and pathological changes were selected.

Following slice selection, \gls{ROI}(s) were manually identified using bounding boxes, placed individually on a slice-by-slice basis. The annotated \gls{ROI}(s) included the medial temporal lobe, parietal cortex, and posterior cingulate areas commonly affected in \gls{AD}. These bounding boxes were used to localize relevant brain regions displaying characteristic structural changes, such as parenchymal atrophy and ventricular widening, and to guide subsequent detailed assessments of atrophy patterns.

Following initial localization, detailed annotations were evaluated using three standardized visual rating scales. \gls{AD} was assessed on coronal T1-weighted slices perpendicular to the hippocampal axis, following the Scheltens scale (0–4), based on hippocampal size, choroid fissure widening, and temporal horn enlargement. \gls{GCA} was evaluated using the Pasquier scale on axial FLAIR images, with attention to sulcal widening and cortical thinning in the frontal, parietal, and temporal lobes. Posterior atrophy was scored using the \gls{Koedam} across sagittal, axial, and coronal planes, focusing on the precuneus, posterior cingulate, and parieto-occipital sulcus. 

For each region, a score was assigned according to the respective scale, along with a brief textual explanation justifying the score based on visual features (e.g., sulcal widening, hippocampal shrinkage, or cortical thinning). Each region was scored independently in both hemispheres. 

Final annotations were established by consensus, requiring agreement from at least two out of the three expert raters to ensure diagnostic reliability and minimize inter-rater variability.

\onecolumn
\section{Ethical Statements}
\subsection{Copyrights}
\label{sec:copyrights}
\subsubsection{Apache License 2.0}
The Apache License, Version 2.0 (Apache 2.0) is a \textbf{permissive open-source license} developed by the Apache Software Foundation (ASF). Its main characteristics are:

\begin{itemize}
    \item \textbf{Free Use:} The software can be used for any purpose, including commercial applications.
    \item \textbf{Modification \& Distribution:} Users may modify the code and redistribute original or modified versions.
    \item \textbf{Attribution:} A copy of the license must be included and proper credit given to the original authors.
    \item \textbf{NOTICE File:} If the project includes a \texttt{NOTICE} file, it must be preserved during redistribution.
    \item \textbf{Patent Grant:} Contributors grant users a license to patents that would otherwise be infringed by their contributions.
    \item \textbf{Disclaimer:} The license provides the software ``as is'' without warranties or liability.
\end{itemize}

\noindent\textbf{Practical Implications:}
\begin{itemize}
    \item Permits integration into proprietary (closed-source) projects.
    \item Allows combination with other open-source or commercial code.
    \item Enables redistribution under new branding.
\end{itemize}

\noindent\textbf{Restrictions:}
\begin{itemize}
    \item License and attribution notices cannot be removed.
    \item Modified versions cannot be misrepresented as the original work.
    \item Original authors cannot be held liable for issues.
\end{itemize}

\subsubsection{Fair Use}

In addition to permissive open-source licenses such as Apache 2.0, the doctrine of \textbf{Fair Use} provides a legal framework that may justify the reuse of third-party datasets for research and educational purposes. Fair Use is codified under United States copyright law (17 U.S.C. \S107) and is widely invoked in academic contexts, such as in our work. Its applicability is assessed through four key factors:

\begin{enumerate}
    \item \textbf{Purpose and character of use:} Non-commercial, educational, and research-driven usage is generally favored. Transformative use---where the dataset is repurposed for new scientific insights rather than replicating its original function---strengthens the case.
    \item \textbf{Nature of the copyrighted work:} Factual and scientific data are afforded less stringent protection compared to creative works, which supports their reuse in research.
    \item \textbf{Amount and substantiality:} Use of limited portions, or selective aspects of the dataset, weighs in favor of Fair Use. However, even large-scale use can be justified if it is essential for the research objective and transformative in nature.
    \item \textbf{Effect on the market:} If the research use does not undermine the commercial market or value of the original dataset, this criterion supports Fair Use.
\end{enumerate}

\textbf{Implications for research:}  
In practice, the reuse of existing datasets is often considered Fair Use when (i) the purpose is non-commercial and scholarly, (ii) the dataset is employed in a novel or transformative manner (e.g., re-annotating, constructing new benchmarks, or deriving insights not intended by the original authors), and (iii) proper attribution is provided. 

While Fair Use is context-dependent and not absolute, adherence to these principles allows researchers to \textbf{legally and ethically justify} their use of external datasets in the advancement of science.

\subsection{Community Use and Research Approval}
This dataset is made available to the research community under the terms and conditions described below.

\noindent \textbf{License Inheritance}: This dataset inherits the Apache License 2.0 from the original source dataset. All provisions, rights, and obligations under the Apache License 2.0 remain applicable to this dataset unless explicitly modified in this appendix.

\noindent \textbf{Restrictions on Redistribution}: Redistribution of this dataset in its original form is not permitted. Redistribution is only allowed if the dataset has been substantially modified or extended, such as through the addition of new annotations, metadata, or derived data that provide additional research value.

Any redistributed version must:
\begin{itemize}
  \setlength{\itemsep}{1pt}      
    \item Clearly document all modifications or additions.
    \item Retain the original license notices and attribution statements.
    \item Include reference to this appendix and the terms herein.
\end{itemize}

\noindent \textbf{Attribution and Citation Requirements}: All users of this dataset must:
\begin{itemize}
  \setlength{\itemsep}{1pt}  
    \item Cite the associated publications describing the dataset.
    \item Provide a link to the official dataset page in all derivative works, software repositories, and academic publications utilizing the dataset.
    \item Acknowledge \textbf{both} the original dataset and our derived dataset in any related dissemination materials.
\end{itemize}

\noindent \textbf{Permitted Use}: The dataset is approved for use in research and educational applications, including but not limited to:
\begin{itemize}
  \setlength{\itemsep}{1pt}  
    \item Academic studies and experiments.
    \item Development and evaluation of algorithms.
    \item Teaching and coursework within educational institutions.
\end{itemize}

\noindent \textbf{Intent}: These conditions are established to encourage open and responsible scientific collaboration, while ensuring that the provenance, quality, and integrity of the dataset are preserved.

\subsection{Usage Considerations}
\vspace{-0.05in}
This dataset is intended \textbf{solely for research and educational purposes}, and users are expected to exercise appropriate caution in its use and interpretation. As this is a medical dataset, it may inherently contain \textbf{human and/or machine annotation errors}, as well as potential biases or inconsistencies that are beyond our current understanding or control.

While we have made every effort to ensure ethical data collection, careful annotation, and responsible release, we acknowledge that \textbf{no dataset can be guaranteed to be entirely free of error or unintended harm}. Our team holds a deep respect for ethical principles and believes strongly in the notion of \textbf{karma and responsibility}---we have done our utmost to act in good faith throughout the creation and curation of this dataset.

Users are therefore \textbf{strictly advised} to:
\begin{itemize}
  \setlength{\itemsep}{1pt}      
    \item Pilot test any models, analyses, or clinical applications derived from this dataset before applying them to real-world or human-related scenarios.
    \item Avoid direct clinical use or decision-making based solely on this dataset without rigorous validation and oversight from qualified professionals.
    \item Acknowledge potential limitations in any derivative work or publication.
\end{itemize}

By using this dataset, users accept that the authors and contributors \textbf{cannot be held responsible for any consequences} arising from misuse, misinterpretation, or overreliance on the data. Ethical stewardship and cautious application remain the user's responsibility.


\end{document}